\documentclass[lettersize,journal]{IEEEtran}

\usepackage[table,xcdraw]{xcolor}
\usepackage{amsmath,amssymb,amsfonts}
\usepackage{graphicx}
\usepackage{textcomp}
\usepackage{tabularx}
\usepackage{makecell}
\usepackage{pgfplots}
\usepackage{multirow}
\usepackage{amssymb}
\usepackage{pifont}
\newcommand{\cmark}{\ding{51}}
\newcommand{\xmark}{\ding{55}}
\usepackage{hyperref}
\usepackage{soul}
\usepackage{subcaption}

\DeclareRobustCommand{\hlpink}[1]{{\sethlcolor{white}\hl{#1}}}

\usepackage{changepage}
\usepackage{pdflscape}
\usepackage{comment} 
\usepackage{multicol}
\usepackage{algorithm}
\usepackage[noend]{algpseudocode}
\usepackage{longtable}
\usepackage{adjustbox}
\usepackage{arydshln}
\usepackage{lipsum}
\usepackage{supertabular}
\usepackage{url}
\usepackage{stfloats}

\newcommand\SmallerCaption[1]{%
  \captionsetup{font=scriptsize}%
  \caption{#1}
 }

\pgfplotsset{compat=1.11,
    /pgfplots/ybar legend/.style={
    /pgfplots/legend image code/.code={%
       \draw[##1,/tikz/.cd,yshift=-0.25em]
        (0cm,0cm) rectangle (3pt,0.8em);},
   },
}

\pgfplotsset{
    compat=1.11,
    legend image code/.code={
    \draw[mark repeat=2,mark phase=2]
    plot coordinates {
        (0cm,0cm)
        (0.1cm,0cm)        %
        (0.2cm,0cm)         %
        };%
    }
}

\def\BibTeX{{\rm B\kern-.05em{\sc i\kern-.025em b}\kern-.08em
    T\kern-.1667em\lower.7ex\hbox{E}\kern-.125emX}}

\algnewcommand{\IfThenElse}[3]{%
  \State \algorithmicif\ #1\ \algorithmicthen\ #2\ \algorithmicelse\ #3}
\algnewcommand{\IfThen}[3]{%
  \State \algorithmicif\ #1\ \algorithmicthen\ #2\ }
\begin{document}
\bstctlcite{IEEEexample:BSTcontrol}

\title{MalProtect: Stateful Defense Against Adversarial Query Attacks in ML-based Malware Detection}

\author{Aqib~Rashid, Jose~Such%
\IEEEcompsocitemizethanks{\IEEEcompsocthanksitem Both authors are with the Department of Informatics, King's College London, Strand, London WC2R 2LS, United Kingdom. Jose Such is also with VRAIN, Universitat Polit\`{e}cnica de Val\`{e}ncia, Spain.\protect\\
E-mail: \{aqib.rashid, jose.such\}@kcl.ac.uk}%
}

\markboth{}%
{MalProtect: Stateful Defense Against Adversarial Query Attacks in ML-based Malware Detection}

\maketitle
 
\begin{abstract}
ML models are known to be vulnerable to adversarial query attacks. In these attacks, queries are iteratively perturbed towards a particular class without any knowledge of the target model besides its output. The prevalence of remotely-hosted ML classification models and Machine-Learning-as-a-Service platforms means that query attacks pose a real threat to the security of these systems. To deal with this, stateful defenses have been proposed to detect query attacks and prevent the generation of adversarial examples by monitoring and analyzing the sequence of queries received by the system. Several stateful defenses have been proposed in recent years. However, these defenses rely solely on similarity or out-of-distribution detection methods that may be effective in other domains. In the malware detection domain, the methods to generate adversarial examples are inherently different, and therefore we find that such detection mechanisms are significantly less effective. Hence, in this paper, we present MalProtect, which is a stateful defense against query attacks in the malware detection domain. MalProtect uses several threat indicators to detect attacks. Our results show that it reduces the evasion rate of adversarial query attacks by 80+\% in Android and Windows malware, across a range of attacker scenarios. In the first evaluation of its kind, we show that MalProtect outperforms prior stateful defenses, especially under the peak adversarial threat. 
\end{abstract}


\vspace{-2mm}

\section{Introduction}
\label{sec:introduction}
\IEEEPARstart{M}L has offered enormous capabilities, leading to the large-scale use of Machine-Learning-as-a-Service (MLaaS) \cite{apruzzese2023realgradients}, which allows users to leverage remotely-hosted ML models by requesting predictions from them \cite{yu2020cloudleak}. The widespread use of these services means that their reliability and security is paramount, especially considering the threat posed by adversarial machine learning \cite{szegedy2013intriguing,papernot2017practical,rosenberg2018generic}. Recent work has shown that these systems are vulnerable to adversarial query attacks (e.g., \cite{brendel2017decision, chen2020hopskipjumpattack, rosenberg2020query}).  %
With query attacks, an attacker iteratively perturbs a malware sample based on feedback from the target model that includes its predicted output. Using the feedback, the perturbations to the input sample are tuned across the queries to observe how the target model responds until, eventually, the truly malicious sample is classified as benign (i.e., goodware) \cite{chen2020hopskipjumpattack, rosenberg2020query}.
To deal with the adversarial ML threat, several defenses have been proposed based on a variety of approaches (e.g., \cite{szegedy2013intriguing,papernot2016distillation, tramer2017ensemble, sengupta2018mtdeep}).
However, most of these approaches have been shown to be ineffective against \emph{query} attacks in several domains \cite{ilyas2018black, brendel2017decision,chen2020hopskipjumpattack}, including ML-based malware detection \cite{rosenberg2018generic,rosenberg2020query,pierazzi2020problemspace}.

Recent work has stressed that systems must monitor queries to identify hazards such as adversarial attacks \mbox{\cite{hendrycks2021unsolved}}. In fact, ML-based anomaly detection has been regarded as essential to detect the misuse of ML-based systems \cite{brundage2018malicious}. 
To this end, \emph{stateful defenses} have been proposed to protect ML prediction models against query attacks \cite{chen2020stateful} by offering greater system awareness. This is achieved by monitoring queries received by the system. Researchers have hypothesized that sequences of adversarial queries are often abnormally similar to each other, unlike sequences of legitimate queries \cite{chen2020stateful, li2020blacklight, juuti2019prada}, and often do not fit their distribution \cite{juuti2019prada, Kariyappa_2020_CVPR, atli2020extraction}. %
Hence, stateful defenses analyze the sequence of queries made to the system, such as the similarity of feature vectors representing an image. 
Several stateful defenses have been proposed (e.g., \cite{chen2020stateful, li2020blacklight, seat, juuti2019prada, pal2021stateful}); however, prior to our work, they have not been tested in the malware detection domain, where query attacks can cripple prediction models %
\cite{rosenberg2018generic, rosenberg2020query}. The malware detection domain is significantly different from other domains, with more constraints imposed on attackers regarding the discrete representation of feature vectors and the preservation of malicious functionality when generating adversarial examples \cite{rosenberg2018generic, biggio2013evasion}. Therefore, attackers use techniques that are different from those in other domains to generate adversarial examples \cite{rosenberg2020query, pierazzi2020problemspace}. Consequently, stateful defenses that %
have been applied to other domains may, in fact, be ineffective when tested in the malware detection domain.%

Hence, in this paper, we present \emph{MalProtect}, which is a model-agnostic stateful defense %
against query attacks in the ML-based malware detection domain. 
\hlpink{MalProtect employs several threat indicators that use different data-driven statistical and ML-based methods to analyze the sequence of queries received by the system. Upon receiving a new query and conducting an analysis, the threat indicators each produce scores reflecting the likelihood of an attack based on different criteria. These are then aggregated to predict if there is an attack in progress. This allows MalProtect to detect attacks in a manner that is reliable and interpretable.} Across Android and Windows, we show that MalProtect reduces the evasion rate of query attacks by 80+\%, and we compare this to the meager performance of prior stateful defenses across a range of attack scenarios in the first evaluation of its kind. %
We further show that, despite being designed to protect against query attacks, MalProtect affords a degree of protection against other types of attacks, such as transferability attacks. Furthermore, we demonstrate that an adaptive attacker --- with complete knowledge of MalProtect --- cannot achieve significant evasion. In summary, we make two key contributions:

\begin{enumerate}
    \item We propose the first stateful defense against adversarial query attacks in ML-based malware detection. Our defense uses several threat indicators to prevent query attacks and the generation of adversarial examples.
    
    \item We provide the first evaluation of several prior stateful defenses applied to the ML-based malware detection domain. We show that MalProtect reduces the evasion rate of query attacks across Android and Windows under a range of threat models and attacker scenarios.

\end{enumerate}

\hlpink{The rest of this paper is organized as follows: \mbox{Section~\ref{sec:background}} provides the background and related work. \mbox{Section~\ref{sec:threatmodel}} defines the threat model. \mbox{Section~\ref{sec:ourmethod}} describes MalProtect. \mbox{Section~\ref{sec:expsetup}} details the experimental setting. In \mbox{Sections~\ref{sec:blackboxattack}-\ref{sec:overhead}}, we present our experimental results. In \mbox{Section~\ref{sec:conclusion}}, we conclude.}

\vspace{-2mm}
\section{Background \& Related Work}
\label{sec:background}
\noindent{\textbf{ML-based Malware Detection.}}
ML-based malware detection has rapidly grown in popularity. It allows for unseen and unknown threats to be discovered and offers performance that surpasses other detection approaches \cite{rosenberg2018generic}. ML-based malware detection classifiers decide if a query is benign (i.e., goodware) or malware. These models are trained on binary feature vectors representing the presence or absence of features in executables \cite{rosenberg2018generic, grosse2017adversarial}. The quality of the predictions is determined by these features, which can include the usage of API calls, the network addresses accessed, or the libraries used \cite{grosse2016, al2018adversarial, rosenberg2018generic, demontis2017yes}. 

\noindent{\textbf{Adversarial ML Attacks.}} However, ML models are vulnerable to adversarial ML attacks. Adversarial examples can be developed by an attacker, which are queries that are designed to evade a classifier. Even without direct access to the target model, an attacker can perturb a malware sample to have it classified as benign and evade the prediction model through \emph{transferability attacks} and \emph{query attacks}. With transferability attacks, an estimation of the target model is developed and attacked in anticipation that the generated adversarial examples will \emph{transfer to} the target model \mbox{\cite{papernot2017practical}}. This is based on the transferability property of adversarial examples \mbox{\cite{szegedy2013intriguing}}.

Meanwhile, query attacks %
generate an adversarial example by iteratively perturbing an input sample towards the desired class based on feedback received from the target model until evasion is achieved \cite{rosenberg2018generic, chen2020hopskipjumpattack}. This attack can be conducted in a black-box manner without any knowledge about the target model besides the predicted outputs. In image recognition, recent work has introduced several query attacks using techniques such as gradient and decision boundary estimation that perturb the continuous feature-space of the domain \cite{brendel2017decision, chen2020hopskipjumpattack, ilyas2018black}. However, in the malware detection domain, the functionality of software executables must be preserved, and discrete feature vectors must be used when generating adversarial examples \cite{grosse2017adversarial, rosenberg2018generic, rosenberg2020query}. For example, a feature representing an API call (e.g., \(CreateFile()\)) cannot be perturbed continuously by an attack (e.g., \(CreateFile() + 0.05\)); rather, a different feature must be used \cite{rosenberg2020query} that offers the same functionality. Therefore, software transplantation-based techniques have been proposed to achieve evasion that use features from benign samples to perturb a malware sample %
\cite{rosenberg2020query, pierazzi2020problemspace}.

\noindent{\textbf{Defenses Against Adversarial ML.}} Many defenses against adversarial ML have been proposed (e.g., \cite{szegedy2013intriguing,papernot2016distillation, tramer2017ensemble,sengupta2018mtdeep}). 
Most approaches are single-model defenses, where an individual model is made robust \cite{pang2019improving}. Ensemble defenses have also been proposed, which use several models with some particular method to return predictions \cite{cheng2020voting,rashid2022stratdef, sengupta2018mtdeep, shahzad2013comparative}. However, recent studies (e.g., \cite{carlini2019evaluating,athalye2018obfuscated, rosenberg2018generic}) have shown that most defenses are ineffective against query attacks. As these defenses are \emph{stateless}, they cannot analyze the sequence of queries received and prevent query attacks.

\begin{figure}[!ht]
\vspace{-5mm}
\centering
    \scalebox{0.8}{
        \includegraphics{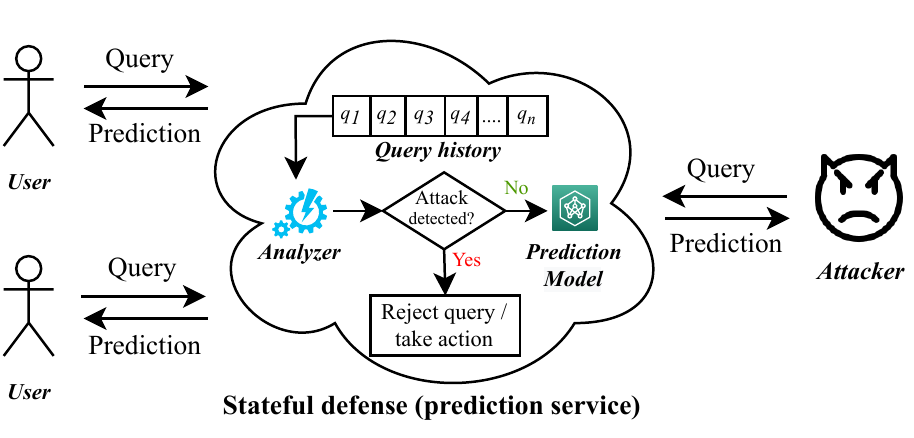}
    }
\vspace{-3.5mm}
    \caption{Overview of a stateful defense.}
  \label{figure:statefuldefence}
\vspace{-3mm}
\end{figure}

\noindent{\textbf{Stateful Defenses.}} To defend against query attacks, \emph{stateful defenses} have been proposed (e.g., \cite{chen2020stateful, li2020blacklight, juuti2019prada,Kariyappa_2020_CVPR, atli2020extraction, 10.1145/3560830.3563727}). Stateful defenses conduct a form of anomaly detection, which has been recommended in recent studies as a method to combat adversarial query attacks \cite{hendrycks2021unsolved,brundage2018malicious,9144212}. It has been hypothesized that because query attacks iteratively perturb an input sample, they produce a sequence of abnormal queries that have high similarity or are outside the distribution of legitimate queries. Figure~\ref{figure:statefuldefence} shows that stateful defenses retain the sequence of queries received by the system and analyze them using different techniques to detect attacks. This is similar to intrusion detection systems and firewalls that monitor network activity to detect threats \cite{10.17487/RFC1636}.  %

\hlpink{However, stateful defenses that protect (remotely-hosted) ML models are different from traditional API protection mechanisms. To protect remotely-hosted services, traditional API protection mechanisms include simple rate-limiting, encryption, and input validation. Conversely, stateful defenses analyze the sequence of queries received by the system by examining their similarity, their distribution, and other factors before queries are passed to the underlying prediction model that is under protection. This means that attacks can be detected before they can cause harm to the ML model. %
}

\noindent{\textbf{Limitations of Existing Stateful Defenses.}} A number of stateful defenses have been proposed in recent years \cite{chen2020stateful, juuti2019prada, Kariyappa_2020_CVPR, seat, li2020blacklight}. 
Subsequent work has identified several limitations of these defenses, including insufficient similarity detection mechanisms, poor maintenance of the query history, and weaknesses of account-based detection, among others \cite{li2020blacklight, pal2021stateful, seat, hendrycks2021unsolved}. 
\hlpink{In \mbox{Table~\ref{table:statefulrelatedlit}}, we provide an overview of stateful defenses, covering the different approaches that have been employed in prior work. Subsequently, we provide detail about each of these approaches, discussing how they detect attacks and any key limitations.}

\begin{table}[!htbp]
\scalebox{0.74}{
\begin{tabular}{lllllll}
\hline
 & \makecell[l]{Chen et al.\\(SD)\\ \cite{chen2020stateful}} & \makecell[l]{Li et al.\\(Blacklight)\\\cite{li2020blacklight}} & \makecell[l]{Azmoodeh\\ et al.\\\cite{10.1145/3560830.3563727}} & \makecell[l]{Kariyappa\\ et al.\\\cite{kariyappa2019improving}} & \makecell[l]{Atli et\\ al.\\\cite{atli2020extraction}} & \makecell[l]{Juuti et al.\\(PRADA)\\\cite{juuti2019prada}}\\
\hline
\makecell[l]{Maintains query\\ history} & \cmark & \cmark & \cmark &  &  & \cmark  \\
\hdashline[0.5pt/5pt]
\makecell[l]{Distance-based /\\ Similarity detection} & \cmark & \cmark & \cmark &  &  & \cmark \\
\hdashline[0.5pt/5pt]
\makecell[l]{Account-based\\ defense} & \cmark &  & \cmark &  &  & \cmark  \\
\hdashline[0.5pt/5pt]
\makecell[l]{Out-of-distribution\\  (OOD) detection} &  &  &  & \cmark & \cmark & \cmark \\
\hdashline[0.5pt/5pt]
\makecell[l]{OOD based on\\ training data} &  &  &  & \cmark & \cmark &  \\ 
\hline
\end{tabular}
}
\caption{\hlpink{Overview of stateful defenses and attack detection for adversarial ML.}}
\label{table:statefulrelatedlit}
\vspace{-2mm}
\end{table}

More specifically, Chen et al.'s Stateful Detection approach (SD) is an account-based defense that maps a user's query to a feature vector and measures the average distance between it and its k-nearest neighbors. If this is below a threshold, an attack is detected, which resets the query history and cancels the user's account. However, this can be defeated by a Sybil attack, where an attacker uses multiple accounts in the attack \cite{10.5555/646334.687813, yang2021policydriven}. Moreover, inspecting queries per-account limits the detection scope \cite{li2020blacklight}. %
Alternatively, Li et al. present Blacklight \cite{li2020blacklight}, which measures the similarity of queries  for all users using the \(L_{2}\) distance (i.e., Euclidean distance), which is a type of \(L_{p}\) norm \cite{carlini2017towards}. An attack is detected if the Euclidean distance between several queries (represented by hashes) is below the threshold. Other distance-based defenses have been proposed (e.g., \cite{10.1145/3560830.3563727}), though these can all be evaded when queries are intentionally designed to be dissimilar \cite{hendrycks2021unsolved} or through query-blinding attacks \cite{seat}.

Out-of-distribution (OOD) detectors have also been proposed to thwart query attacks and detect adversarial examples \cite{9144212}. These check whether a query belongs to the distribution of the target model's training data \cite{9144212, Kariyappa_2020_CVPR, atli2020extraction}. When OOD queries are detected, Kariyappa et al. propose returning inconsistent labels to those detected OOD queries \cite{Kariyappa_2020_CVPR}. However, an attacker may be able to construct adversarial queries that remain within the required distribution \cite{seat,yu2020cloudleak}. %
PRADA \cite{juuti2019prada} is a defense against model extraction attacks, which is a related problem. Prior work has found that it can also detect evasion attacks \cite{chen2020stateful}. PRADA is based on the assumption that the \(L_{2}\) distance among non-adversarial queries follows a normal distribution. This is monitored using the Shapiro-Wilk normality test. However, manipulation of the query distribution has been found to evade it \cite{chen2020stateful, yu2020cloudleak}. %

Our defense, MalProtect, is model-agnostic allowing it to be used with any underlying prediction model. %
To detect attacks, it utilizes techniques beyond basic similarity and OOD detection, such as analyzing the autoencoder loss of queries, the distribution of enabled and shared features across queries, and more. %
We next present the threat model we consider for our work, followed by a detailed description of MalProtect.

\vspace{-3mm}

\section{Threat Model}
\label{sec:threatmodel}
\vspace{-1mm}
In our work, attackers aim to evade a feature-based ML malware detection classifier so that their malicious queries are misclassified as benign. This is a well-established threat model in this domain \cite{severi2021explanation, yang2017malware, suciu2019exploring, grosse2016, grosse2017adversarial, biggio2013evasion}.

\noindent{\textbf{Target Model.}}
The target model is a remotely-hosted malware classification system that predicts whether an input sample belongs to the benign or malware class. We refer to the classification system as the oracle \(O\). The oracle has two principle components: the underlying classifier \(F\) (i.e., the prediction model) \hlpink{that predicts whether an input sample is malicious} and the stateful defense that protects \(F\) by analyzing queries to detect attacks (e.g., MalProtect --- see Section~\ref{sec:ourmethod} later). \hlpink{In our work, we use a range of prediction models from prior work (see \mbox{Section~\ref{sec:expsetup}} later).} \hlpink{Generally, to train such a prediction model} \(F\), input samples are represented as binary feature vectors using the features that are provided by a dataset. With the features \(1...M\), a feature vector \(X\) can be constructed for each input sample such that \(X \in \{{0,1}\}^M \), similar to previous work \cite{rashid2022stratdef, pods, grosse2017adversarial}. The presence or absence of a feature \(i\) is represented by either \(1\) or \(0\) within \(X\). Like prior work on ML-based malware detection \cite{rosenberg2020query, grosse2017adversarial, li2021framework}, the features we employ in our work include, among others, libraries, API calls, permissions, and network addresses; these are provided by the datasets we use (see Section~\ref{sec:expsetup} later). With their class labels, several feature vectors can be used to train the binary classification model that we refer to as the classifier \(F\). 
Then, when a user makes a query to \(O\), a prediction is made and returned. 
For the predicted outputs, we use 1 for the malware class and 0 for the benign class. \(O\) returns scoreless feedback \cite{ilyas2018black}. %
\hlpink{Recall that $O$ consists of the prediction model and the independent stateful defense designed to protect it.}

\setlength\abovedisplayskip{0pt}%
\setlength\belowdisplayskip{8pt}%
\setlength\abovedisplayshortskip{-8pt}%
\setlength\belowdisplayshortskip{2pt}%

\vspace{1mm}
\noindent{\textbf{Attacker's Goal.}}
The goal of the attacker is to generate an adversarial example \(X'\) from a malware sample \(X\) to evade the oracle \(O\) and obtain a benign prediction. Suppose \(O\colon X \in \{{0,1}\}^M \) and we have a function \(check()\) to check the \hlpink{malicious} functionality of \(X\). We can summarize this as:
\begin{align}
    check(X) = check(X'); O(X) = 1; O(X') = 0 
\end{align}

\hlpink{Other variations of this goal do exist (e.g., achieving evasion with minimal queries \mbox{\cite{10.1145/3433667.3433669}}). However, we focus on the more general goal above, as it is commonly explored in related work on stateful defenses and ML-based malware detection.}

\noindent{\textbf{Attacker Capabilities \& Knowledge.}} 
\hlpink{In order to evaluate MalProtect and other stateful defenses under different attack scenarios and conditions,} we model three types of attackers that have been featured in prior work \cite{ilyas2018black, laskov2014practical, papernot2018sok, biggio2013evasion} \hlpink{as is common in prior work}. We consider the \emph{gray-box attacker}, who has limited knowledge, has access to the same training data as the classifier \(F\) and is aware of the feature representation and the statistical representation of the features across the dataset. However, the attacker is unaware of \(F\)'s parameters, configurations, and other pertinent information. This represents a scenario similar to when some model data has been leaked. To perform query attacks, this attacker applies perturbations to the feature vector through a software transplantation-based approach \cite{rosenberg2020query, pierazzi2020problemspace}, guided by their knowledge of the dataset. In contrast, the zero-knowledge \emph{black-box attacker} can only observe the predictions received for their queries and has no information about the target model but does have some information pertaining to the kind of feature extraction performed (e.g., the static analysis that a malware detection classifier may consider). This attacker also uses a software transplantation-based approach, but with considerably less information. Neither attacker is aware that MalProtect is a part of \(O\). We also consider an \emph{adaptive attacker} (i.e., a \emph{white-box attacker}) who %
has complete knowledge of the target model and knows that MalProtect is in place. Therefore, this attacker tries to evade all components of the oracle \(O\) with a specific, tailored attack.

We assume that all attackers possess ample computing power and resources to submit thousands of queries. Furthermore, unlike previous work \cite{juuti2019prada, chen2020stateful}, we assume that attackers may  control multiple user accounts and IP addresses, as in a Sybil attack %
\cite{10.5555/646334.687813, yang2021policydriven}. \hlpink{As attackers can submit many queries, they may be able to conduct a transferability attack by constructing substitute models \mbox{\cite{papernot2017practical}}, so that a black-box attack becomes a semi white-box one. However, we mainly focus on query attacks in this work, as stateful defenses are designed to primarily defend against them \mbox{\cite{chen2020stateful}}. Moreover, were MalProtect to be deployed in a production environment, it would be unreasonable to assume that basic rate-limiting protections were not in place to prevent excessive queries by users, making such transferability attacks less practical.}

\vspace{-2mm}

\section{MalProtect}
\label{sec:ourmethod}

\begin{figure}[!htbp]
\vspace{-3mm}
    \scalebox{1.35}{
        \includegraphics{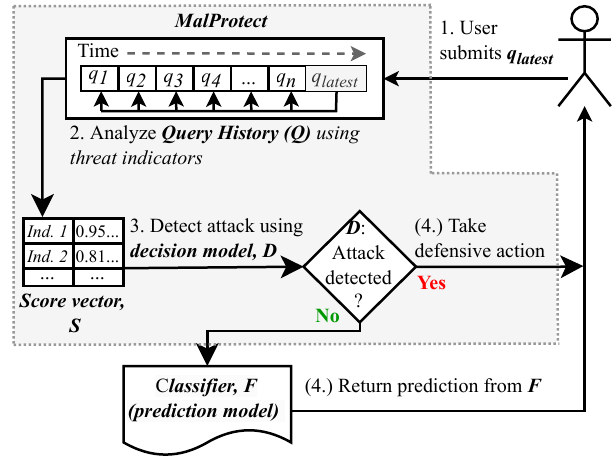}
    }
    \caption{Overview of MalProtect.}
  \label{figure:}
\vspace{-4mm}
\end{figure}

MalProtect is designed to protect ML models against adversarial query attacks in the malware detection domain's feature-space. %
Figure~\ref{figure:} shows that MalProtect forms a layer of protection that monitors queries before they reach the prediction model $F$, which is the ML-based malware detection classifier to be protected. This allows MalProtect to detect adversarial queries that may evade the classifier (i.e., prediction model).  The prediction model $F$ can be a vanilla ML model or a defense (e.g., an adversarially-trained model, an MTD, etc.). By having a defense as the prediction model used with MalProtect, it can offer \emph{defense in depth}, as multiple layers of security can contribute to the overall security of the system.

When a user submits a query (Step 1), MalProtect employs several \emph{threat indicators} to analyze the sequence of queries for attack detection (Step 2). This is similar to a system that makes decisions using data from multiple sensors \cite{hendrycks2021unsolved}. The threat indicators use different criteria, principally driven by an anomaly detection-based approach. This includes examining aspects such as the autoencoder loss of queries, the distance between queries, the features shared across queries, and whether other characteristics of queries fit within different distributions. 
Based on the state of the query history, each indicator produces a score that reflects the likelihood of an attack in progress according to that indicator, with adversarial queries expected to cause higher scores. %
A \emph{decision model} then takes the indicator scores as its input and predicts whether there is an attack in progress (Step 3). If an attack is detected, MalProtect takes some defensive action. If no attack is detected, the prediction model $F$ returns a prediction for the user's query (Step 4). Not only do the scores assist in interpreting how MalProtect produces a particular decision, but as we show later, each threat indicator can be analyzed to determine its influence on the final decision. Next, we provide details about each of MalProtect's core components.

\vspace{-4mm}

\subsection{Query History}
MalProtect analyzes the query history to detect adversarial behavior. When a user requests a prediction for their query \(q_{latest}\), it is appended to the query history \(Q\). Queries can be represented more sparsely in the malware detection domain to reduce storage overhead, which has been cited as an issue in other domains \cite{li2020blacklight}. 
The depth of analysis performed by MalProtect and the size of \(Q\) are controlled by the resources possessed by the defender when MalProtect is deployed, but, as we show later in Section~\ref{sec:overhead}, MalProtect needs very little spatial and computational resources to offer a very good performance. Importantly, the size of $Q$ will obviously never be infinite \cite{miller2016reviewer, 244706}, though it can extend as far as the defender's resources allow. %
Unlike other stateful defenses that reset the query history once they reach capacity (e.g., \cite{chen2020stateful}), MalProtect retains queries as a sliding window to maximize its detection capability. %
Finally, MalProtect does not distinguish between queries from different users, as an attacker can conduct a Sybil attack to evade account-based defenses \cite{10.5555/646334.687813, yang2021policydriven}. %
\vspace{-3mm}

\subsection{Indicators for Analyzing Query History}
After adding \(q_{latest}\) to \(Q\), MalProtect analyzes the query history using multiple \emph{threat indicators} to detect attacks through an anomaly detection-based approach. In conventional anomaly detection, thresholds are employed to assess whether a sample meets the criteria for being abnormal. However, with MalProtect, we use a data-driven approach where the weights of each indicator are learned by a \emph{decision model} to predict an attack. That is, instead of each indicator deciding if an attack is taking place, each indicator produces a score between 0 and 1 that reflects the likelihood of an attack according to its criteria. Scores are then aggregated and passed to a decision model that decides if an attack is in progress. This allows for a degree of interpretability too, as we can understand which indicators play the biggest role in the decision model and which indicators produce the highest scores (giving clues about what is anomalous), as we show later. %
MalProtect is based on the principle that adversarial queries are expected to cause higher indicator scores as they are anomalous compared to legitimate queries and the training data distribution.

To conduct the analysis and produce scores, some indicators use standard anomaly detection techniques, such as the empirical rule \cite{10.2307/2684253}, to assess whether a query fits the distribution of data. Meanwhile, other indicators use information about the training data to examine how much queries deviate from various measures. We next present a detailed description of each indicator:

\noindent{\textbf{\textit{Indicator 1. Distance between queries:}}} This indicator assesses the similarity of \(q_{latest}\) to other queries in \(Q\), which is the primary method adopted by some stateful defenses (e.g., \cite{chen2020stateful, li2020blacklight}). A smaller distance between queries is an indication of iterative perturbations during a query attack, as an attacker makes modifications to their queries to assess how the target model responds \cite{chen2020stateful,chen2020hopskipjumpattack}. 

This indicator therefore examines how anomalous the minimum distance between \(q_{latest}\) and other queries in \(Q\) is (represented by \(minDistQ\)). 
As the malware detection domain uses binary feature vectors (unlike other domains, e.g., image recognition), the $L_{0}$ distance is the most appropriate distance metric \cite{carlini2017towards, pods}, as it measures the number of features that are different between two feature vectors. Other norms, such as \(L_{\infty}\) would always have a distance of 1 if at least one change was present between two feature vectors \cite{pods}. %
To assess the potential abnormality of \(minDistQ\) without using a threshold and also to normalize it between 0 and 1, we put \(minDistQ\) in the context of the average \(L_{0}\) distance of all samples in the training data (represented by \(avgDistD\)). %
If \(minDistQ\) is significantly below this, it implies that \(q_{latest}\) is similar to another query in \(Q\) beyond the norm. 

The score for this indicator is based on the percentage change between \(minDistQ\) and \(avgDistD\). A smaller percentage change means that there exists a query whose similarity to \(q_{latest}\) is significantly lower than the average of the training data (\(avgDistD\)). So that a higher threat is represented by a higher score value, we negate the result. Thus, the score for this indicator is calculated as follows:
\begin{equation}
    S_{1} = -1 \cdot (minDistQ - avgDistD) \mathbin{/} avgDistD
\end{equation}

\noindent{\textbf{\textit{Indicator 2. Enabled features shared across queries:}}} This indicator assesses whether \(q_{latest}\) shares a significant number of enabled features with another query in \(Q\) beyond the norm. The rationale behind this \hlpink{indicator} is that an attacker could enable unused features in \(q_{latest}\) to intentionally increase the \(L_{0}\) distance between queries in order to evade basic similarity detection (e.g., while trying to evade Indicator 1). \hlpink{This is an attack strategy specific to the malware detection domain. That is, the query distance can be increased (e.g., by adding features in bulk) without disrupting the original malicious functionality (e.g., through dead-code, opaque predicates). Conversely, in domains such as image recognition, this attack strategy would not be as useful since adding features to deliberately increase the query distance in this manner would lead to the original image being distorted.}

Intriguingly, though, some \hlpink{core features of malicious queries} must \emph{always} remain enabled to preserve the original malicious functionality across a query attack. Thus, queries with the same enabled features \hlpink{(and therefore shared)} would imply similarity specifically in this domain. Therefore, the highest number of enabled features shared between \(q_{latest}\) and another query in \(Q\) is calculated as \(maxSharedQ\).

We then use the average number of enabled features shared between samples in the training data (\(avgSharedD\)) to assess to what extent \(maxSharedQ\) may be abnormal. This is achieved by calculating the deviation of \(maxSharedQ\) from this mean as a percentage change. A higher value indicates that \(q_{latest}\) shares an abnormally high number of enabled features with some other query in \(Q\) considering the training data's distribution. The score is calculated as follows:
\begin{equation}
    S_{2} = (maxSharedQ - avgSharedD) \mathbin{/} avgSharedD
\end{equation}

\noindent{\textbf{\textit{Indicators 3A \& 3B. Number of enabled features:}}} An attacker could also rapidly traverse the decision boundary by enabling a substantial number of benign features in \(q_{latest}\) at once. \hlpink{That is, by adding many features to a malware sample that are present in benign executables, a misclassification can be achieved. Therefore,} an anomalous number of enabled features could indicate an attack attempt. Hence, the Indicator 3A %
is the percentage change between the number of enabled features in \(q_{latest}\) (\(|q_{latest}|\)) and the average number of enabled features in training data samples (\(avgFeaturesD\)). A large percentage change implies that far more features are enabled in \(q_{latest}\) compared to \(avgFeaturesD\). The score is calculated as follows:
\begin{equation}
    S_{3A} = (|q_{latest}| - avgFeaturesD) \mathbin{/} avgFeaturesD
\end{equation}

However, the Indicator 3A only assesses whether \(|q_{latest}|\) is anomalous considering the training data and does not compare with other queries in \(Q\). It is also useful to assess whether \(|q_{latest}|\) is anomalous considering the distribution of \(Q\) regardless of the training data, similar to previous works (e.g., \cite{Kariyappa_2020_CVPR, atli2020extraction, juuti2019prada}), as most queries are expected to be legitimate. 

Therefore, we also use the Indicator 3B, which assesses to what extent the number of enabled features in \(q_{latest}\) is anomalous considering the queries in \(Q\). The Indicator 3B uses the empirical rule (i.e., 3-\(\sigma\) rule) \cite{10.2307/2684253}, which is a standard technique in anomaly detection. %
The score is calculated as the percentage change from 3 standard deviations of the mean using the following equation, where \(\mu_{FeaturesQ}\) is the mean number of enabled features of queries in \(Q\) and \(\sigma_{FeaturesQ}\) is the standard deviation:
\begin{align}
    C &= \mu_{FeaturesQ} + (\sigma_{FeaturesQ} \times 3)\\
    S_{3B} &= (|q_{latest}| - C) \mathbin{/} C
\end{align}

\noindent{\textbf{\textit{Indicators 4A \& 4B. Distribution similarity via autoencoder reconstruction loss:}}} An autoencoder is a neural network where the input and output are the same but the hidden layers have fewer neurons \cite{schmidhuber2015deep}. This limits the amount of information that can travel through the model, requiring it to learn a compressed version of the input. Interestingly, input samples that belong to the distribution of the autoencoder's training data will produce a smaller distance between the input and output representations (known as the \emph{reconstruction loss}) \cite{10.1145/3439950}. Thus, we hypothesize that an autoencoder trained on legitimate queries will produce a significantly higher reconstruction loss for an adversarial query. Other novelty detection techniques, such as one-class SVMs, have been found to be outperformed by autoencoders in this task in other domains \cite{gong2019memorizing}. 

Hence, Indicator 4A's score is the percentage change between the reconstruction loss of \(q_{latest}\) (\(RecLoss_{q_{latest}}\)) and the maximum reconstruction loss observed in training data samples (\(maxRecLossD\)). A higher percentage change suggests a significant increase from the maximum reconstruction loss observed for training data samples, and that \(q_{latest}\) may be adversarial. The score is calculated as follows:
\begin{equation}
    S_{4A} = (RecLoss_{q_{latest}} - maxRecLossD) \mathbin{/} maxRecLossD
\end{equation}

As with Indicator 3B, we also consider the case where \(q_{latest}\) is only compared to other queries in \(Q\) regardless of the training data distribution. 
In this way, the Indicator 4B uses the empirical rule to assess whether the reconstruction loss of \(q_{latest}\) is anomalous considering the queries in \(Q\). The score is calculated as a percentage change from 3 standard deviations using the following equation, where \(\mu_{RecLossQ}\) is the mean reconstruction loss of queries in \(Q\) and \(\sigma_{RecLossQ}\) is the standard deviation:
\begin{align}
    C &= \mu_{RecLossQ} + (\sigma_{RecLossQ} \times 3)\\
    S_{4B} &= (RecLoss_{q_{latest}} - C) \mathbin{/} C
\end{align}

\vspace{-5mm}

\subsection{Attack Detection}
Once the scores are produced by each indicator, they are aggregated into the score vector \(S\). %
The next step is to make a final decision using all the scores. For this purpose, static techniques can be used to combine several individual scores into an overall score or decision, such as a weighted sum model or a sum of squares model  \cite{doi:10.1287/opre.15.3.537}. However, a key drawback of these static methods is that the weights for each indicator (i.e., how much each indicator contributes to the overall score) must be selected manually by the defender. Moreover, static aggregation models cannot detect any underlying trends or patterns in the data. 

Therefore, MalProtect instead uses a \hlpink{pre-trained} \emph{decision model} (e.g., a neural network) that makes a prediction on the score vector \(S\). Each threat indicator is a feature of this model, with the predicted output from the decision model representing whether there is an attack in progress constituting its decision. \hlpink{Later, in \mbox{Section~\ref{sec:expsetup}}, we provide details about how the decision models can be constructed, including the ones we use in this work (see ``MalProtect Configurations'').} Importantly, MalProtect allows analysts to understand why it has made a particular decision with its scoring system. Analysts can \emph{further} evaluate the influence of each threat indicator on the final prediction by examining the global feature importance of the decision model. Later, in Section~\ref{sec:expsetup}, we present the decision models that we evaluated, and we then discuss how each threat indicator influences the predictions made by these models in Section~\ref{sec:interp}.
\vspace{-5mm}

\subsection{Defensive Action \&  Prediction}
If no attack is detected by the decision model, \(q_{latest}\) is passed on to the classifier \(F\) (also known as the \emph{prediction model}). The classifier makes a prediction of whether \(q_{latest}\) is benign or malware, which is then returned to the user. 
That is, if the latest query received is not considered to be part of an attack by MalProtect, the query is then forwarded to the ML classifier trained to decide whether an input sample is benign or malware. Note that the classifier can itself be made more robust using techniques such as adversarial training (as shown later). MalProtect only acts as a filter, which only lets what it considers to be \emph{legitimate} queries to be passed on to the ML model used for malware classification.

If MalProtect does detect an attack, a defensive action can be taken, such as returning a  specific prediction, returning inconsistent labels, notifying system administrators, banning user accounts, or rejecting further user queries. In the configuration of MalProtect that we use in this paper, if an attack is detected, we take the defensive action of returning a \emph{malware} prediction. This is as if the query had actually been forwarded to the classifier and that had been its output. This means that an attacker would encounter failure throughout the course of the attack, believing that their query is being consistently classified as malware without necessarily knowing about MalProtect's presence. An exciting line of future work would be to study what defensive actions are more or less effective depending on the setting, but this is out of the scope of this paper.%

\vspace{-2mm}

\section{Experimental Setup \& Preliminaries}
\label{sec:expsetup}

\noindent{\textbf{Datasets.}} Sampling from the true distribution is a challenging and open problem in many ML-based security applications \cite{arp2022and,8949524}. Particularly, the lack of publicly accessible, up-to-date datasets in the malware detection domain is a well-known issue that limits the remits of academic research in this domain. To mitigate this, we use three datasets representative of different architectures as well as collection dates and methods that have been used in a variety of studies (e.g., \mbox{\cite{wangrobust, grosse2017adversarial, 10.1145/3503463, pierazzi2020problemspace, 203684, barbero2022transcending, 10.5555/3361338.3361389}}). The datasets we use are AndroZoo for Android malware \cite{Allix:2016:ACM:2901739.2903508}, SLEIPNIR for Windows malware \cite{al2018adversarial}, and DREBIN for Android malware \cite{arp2014drebin}. The consistency of our results across the datasets later show that MalProtect's performance transcends the characteristics and nuances of the datasets.

The AndroZoo dataset \cite{Allix:2016:ACM:2901739.2903508} contains Android apps from 2017 to 2018, offering apps from different stores and markets with VirusTotal summary reports for each. Similar to prior work \cite{pierazzi2020problemspace, 10.5555/3361338.3361389, miller2016reviewer}, we consider an app malicious if it has 4 or more VirusTotal detections, and benign if it has 0 VirusTotal detections (with apps that have 1-3 detections discarded). The dataset contains \(\approx\) 150K recent applications, with 135,859 benign and 15,778 malicious samples. To balance the dataset, we use 15,778 samples from each class. Meanwhile, SLEIPNIR consists of 19,696 benign and 34,994 malware samples. This dataset is derived from Windows API calls in PE files parsed by LIEF. As our work is in the feature-space, we use SLEIPNIR as a representation of Windows out of simplicity due to its feature space, which is binary. This permits a more precise comparison between the Android and Windows datasets. We use 19,696 samples from each class. Meanwhile, DREBIN is based on extracted static features from Android APK files. The dataset contains 123,453 benign and 5,560 malware samples, from which we use 5,560 samples from each class. As in prior work \mbox{\cite{demontis2017yes, grosse2017adversarial}}, and for completeness, we use a large number of features for each dataset, i.e., 10,000 for AndroZoo, 22,761 for SLEIPNIR, and 58,975 for DREBIN. 

Initially, the datasets are partitioned with an 80:20 ratio for training and test data according to the Pareto principle. This training data is partitioned again into training and validation data using the same ratio, producing a 64:16:20 split that has been extensively used in prior work (e.g., \cite{ma2021partner, li2019few}). We use the training data to construct the prediction models used in our evaluation. Meanwhile, we use a partition of randomly-chosen malware samples from a subset of the test data as the input samples for the query attacks, with 234 samples for AndroZoo, 230 for SLEIPNIR, and 229 for DREBIN. 

We also consider the well-established guidelines for conducting malware experiments \cite{6234405}. As our prediction models decide whether an input sample is benign or malicious, it is necessary to retain benign samples in the datasets. Furthermore, we do not strictly balance datasets over malware families but instead over the positive and negative classes. We randomly sample unique samples from each class to appear in the training and test data without repetition \cite{pods, al2018adversarial, rosenberg2018generic}.

\noindent{\textbf{Prediction Models (Non-Stateful Defenses).}} \hlpink{Recall that stateful defenses themselves do not provide predictions, but rather they protect ML prediction models that decide whether an input sample is benign or malicious. Hence, for our experimental evaluation, a selection of prediction models is required.} In our evaluation, we use six \emph{non-stateful} defenses as the prediction models. Rather than using vanilla models, we use single-model defenses, ensemble defenses, and moving target defenses as the prediction models to demonstrate their vulnerability to attacks despite being defenses in their own right. By using MalProtect \hlpink{and other stateful defenses} in conjunction with a defense as a prediction model (rather than a vanilla model), it can introduce defense in depth, which is a frequently-adopted strategy in information security.

\hlpink{We use prediction models from prior work and train them on the three datasets according to the procedures outlined earlier.} Each prediction model is constructed according to the procedures outlined in their original papers using the training data for each dataset (see Appendix~\ref{appendix:modelarchs} for information about architectures). For single-model defenses, we test a neural network with defensive distillation (NN-DD) \cite{papernot2016distillation}. We use several white-box attacks to develop adversarial examples for a set of vanilla models (see later subsection), which are used to adversarially-train a neural network (NN-AT). We adversarially-train with a quantity of adversarial examples that is 25\% of the size of the training data \cite{tramer2017ensemble}. For ensemble defenses, we test majority voting and veto voting \cite{rashid2022stratdef, cheng2020voting, shahzad2013comparative, yerima2018droidfusion, rashid2022stratdef}. For moving target defenses, we evaluate Morphence \cite{amich2021morphence} and  StratDef \cite{rashid2022stratdef}. %
Each prediction model is evaluated against attacks as is (i.e., a non-stateful setup). Then, each prediction model is combined with each stateful defense \hlpink{(e.g., MalProtect)}, allowing us to evaluate different setups and configurations. \hlpink{We next provide details about the MalProtect configurations we use and the other stateful defenses.}

\noindent{\textbf{MalProtect Configurations.}}
We develop two MalProtect configurations to showcase our defense's capabilities. Each configuration uses a different decision model to predict an attack based on the indicator scores. For both configurations, we cap the query history size at 10,000. Later in \mbox{Section~\ref{sec:overhead}}, we evaluate and discuss the \hlpink{efficiency} related to this.

For each configuration, we follow the steps described in \mbox{Section~\ref{sec:ourmethod}} for each dataset. For some indicators, we derive certain values from the dataset that are needed to assess queries (e.g., \(avgDistD\), which is the average \(L_{0}\) distance of samples in the training data). %
Then, to train each decision model \hlpink{(which are used to predict an attack given a score vector)}, a defender could use real-world attack and score data. However, in its absence, a synthetic dataset can be used. Therefore, \hlpink{in this work}, we develop a synthetic dataset of 1000+ labelled samples. For this, we initialize the query history to simulate past user activity with a randomly-chosen set of input samples from the training data. We then supply a range of queries (both legitimate and adversarial, as part of attacks) from our training data to have each indicator produce scores. This produces the synthetic dataset, where each threat indicator is a feature and the class labels represent the true condition of the system at the time (whether it was under attack (1) or not (0)). To predict if an attack is in progress when given a score vector, the decision models we train are a \emph{logistic regression model} for one configuration (MalProtect-LR) and a \emph{neural network} for another (MalProtect-NN) (see Appendix~\ref{appendix:configurations} for architectures). \hlpink{We use an 80:20 split for the training of decision models.} In theory, both models should offer accurate predictions \cite{dreiseitl2002logistic}. \hlpink{When deployed, the decision model may need regular maintenance to ensure that it performs well given the ever-changing threat landscape (see \mbox{Section~\ref{sec:conclusion}} later).}

\noindent{\textbf{Other Stateful Defenses.}}
We compare MalProtect with three stateful defenses that have been applied in other domains to detect query attacks. The \(L_{0}\) defense is based on similarity detection using \(L_{p}\) norms (e.g., \cite{li2020blacklight}); PRADA \cite{juuti2019prada}; and Stateful Detection (SD) \cite{chen2020stateful}. We provide the technical details of each stateful defense in Appendix~\ref{appendix:otherstatefuldefenses}. Although we keep the implementations of these stateful defenses as close to the original as possible (e.g., procedures and parameters), techniques applicable to other domains (e.g., encoding queries to generate a feature vector) are not applicable to the malware detection domain. \hlpink{Recall that stateful defenses do not provide the predictions but are designed to protect the underlying prediction models.}

\noindent{\textbf{Performing Query Attacks.}} \hlpink{To evaluate MalProtect and the other defenses, we apply query attacks under the black-box, gray-box and adaptive attacker scenarios.} When \hlpink{conducting attacks and} generating an adversarial example in the malware detection domain, %
the original malicious functionality must remain intact, and the feature vectors must remain discrete (i.e., consist of 0s and 1s) as they represent the presence or absence of each feature. Query attacks designed for other domains (e.g., \mbox{\cite{chen2020hopskipjumpattack, chen2017zoo}}) do not consider these constraints and cannot produce effective adversarial examples in our domain, as we show in an experiment in \mbox{Appendix~\ref{appendix:additionalqueryattacks}}.

For our domain, query attacks can use software transplantation-based techniques to perturb the features of a malware sample using features from benign ``donor'' samples \hlpink{(which are the benign samples from the test set)}. %
In our evaluation, we modify query attack strategies that have been proven successful in this domain %
\cite{rosenberg2020query}. \hlpink{We refer the reader to  \mbox{Appendix~\ref{appendix:queryattackalgorithm}} for full details of each attack strategy that we use in each scenario. Commonly under each strategy,} we initialize the query history to simulate past user activity (as explained before) prior to starting an attack. Then, using malware samples from our test set, we apply each attack strategy accordingly under each threat model to perturb malware feature vectors. We use the parameter \(n_{max}\) to govern the maximum number of allowed queries permitted, where the transplantation of features continues until the target model is evaded, \(n_{max}\) is reached, or the donor features are exhausted. We apply a procedure to provide a lower bound of  functionality preservation of the original malware sample. %
For this, only valid perturbations for each dataset are allowed %
\mbox{\cite{li2021framework, 8171381, li2020enhancing, al2018adversarial, pierazzi2020problemspace, rashid2022stratdef}}. If a feature has been modified invalidly --- that is, the functionality would not be preserved in the problem-space --- it is restored to its original value. This is to offer a lower bound of functionality preservation within the feature-space, similar to prior work \mbox{\cite{severi2021explanation, grosse2017adversarial, li2021framework}}, in which it was shown that 60+\% of adversarial examples generated following this approach executed as expected in a sandbox environment Recent work \mbox{\cite{li2021framework}}. %

AndroZoo and DREBIN permit feature addition and removal (see Appendix~\ref{appendix:drebinpermitted}). %
For SLEIPNIR, only feature addition is possible because of the processing performed by LIEF to extract features when originally developing the dataset. %
While we remain in the feature-space (like recent work e.g., \mbox{\cite{li2021framework, severi2021explanation}}), the perturbations could be translated to the problem-space using techniques from previous work (e.g., \cite{pierazzi2020problemspace, rosenberg2020query}). For example, feature addition can be achieved by adding dead-code or by using opaque predicates \cite{moser2007limits,pierazzi2020problemspace}. Feature removal --- which is more complex --- can be performed by rewriting the dexcode, encrypting API calls and network addresses (removing features but retaining functionality).

\begin{figure*}[!b]
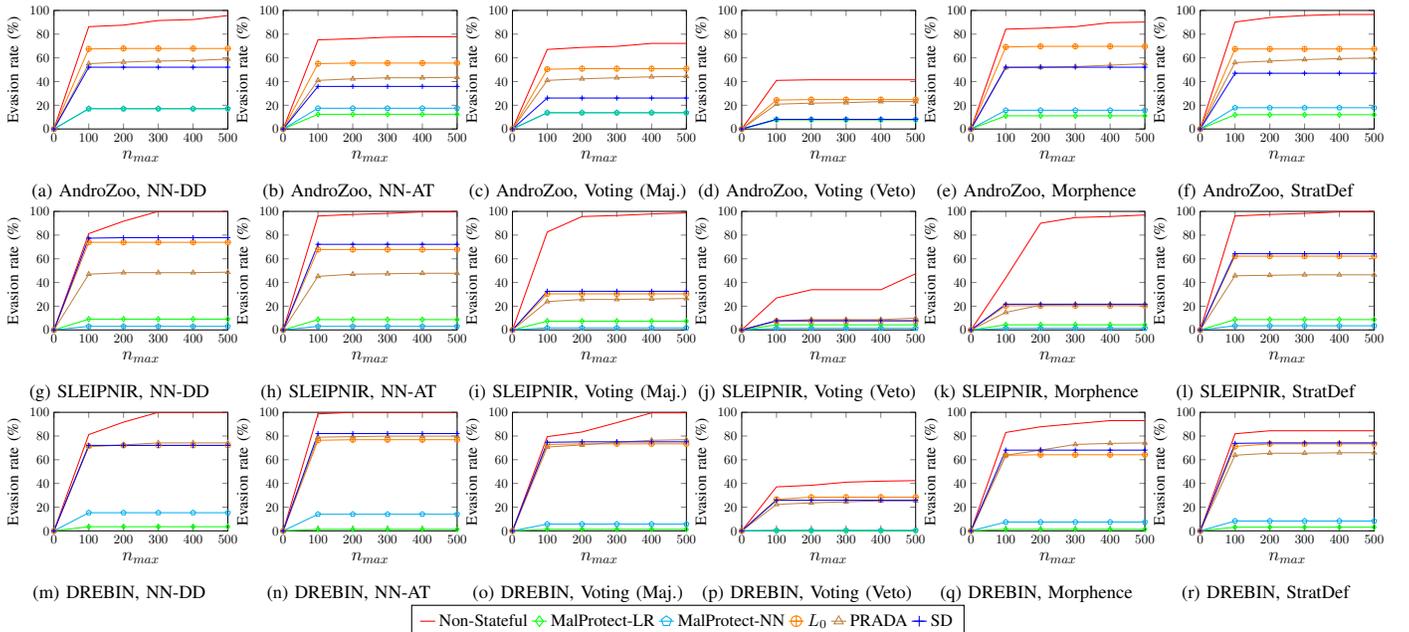

\vspace{-4mm}
	\centering
	 \advance\leftskip-0.2cm
	 
		\begin{subfigure}[b]{0.17\textwidth}

			\SmallerCaption{DREBIN, StratDef}
		\end{subfigure}

		\ref{BlackBoxFullLegend}
\vspace{-2mm}
		\caption{Evasion rate vs. \hlpink{maximum number of queries} (\(n_{max}\)) of black-box query attack against non-stateful defenses (prediction models) and each combination of prediction model and stateful defense.}%
		\label{figure:BlackBoxFull}
		
\vspace{-4mm}
	\end{figure*}

\noindent{\textbf{Performing Other Attacks.}} Prior work \mbox{\cite{li2020blacklight, chen2020stateful, seat}} has used \emph{query blinding attacks} in image recognition with domain-specific techniques (e.g., modifying contrast or brightness). Such techniques are not applicable to our domain. Instead, we later present an adaptive attack strategy, where the attacker has full knowledge of how MalProtect operates, and aims to evade both it and the underlying classifier.

We also use a \emph{transferability attack strategy} to generate adversarial examples \hlpink{using substitute models}. \hlpink{The adversarial examples} are used to adversarially-train models as a defender; and to evaluate stateful defenses later under different system conditions \hlpink{under a transferability attack (in \mbox{Section~\ref{sec:furtherfindings}})}. For this, we construct four vanilla models using the training data (see \mbox{Appendix~\ref{appendix:vanilamodels}} for architectures). \hlpink{These models act as estimations of the prediction models, for they share the same training data, but not the same architectures.} We apply a range of white-box attacks (BIM \cite{kurakin2016adversarial}, Decision Tree attack \cite{grosse2017statistical}, FGSM \cite{goodfellow2014explaining, dhaliwal2018gradient}, JSMA \cite{papernot2016limitations} and SVM attack \cite{grosse2017statistical}) against these models to generate adversarial examples, in anticipation that they transfer to the target models. However, these attacks perturb features without considering the constraints of this domain. Therefore, we preserve the functionality of the original malware sample in the feature-space and discretize the generated feature vectors. That is, once an adversarial example is produced, each value in the feature vector is discretized (i.e., if the value in the feature vector is $< 0.5$, it is set to 0, else it is set to 1). Then, if a feature has been modified invalidly according to functionality constraints, it is restored to its original value (similar to the process for query attacks). We then ensure that the adversarial example can still achieve evasion; if not, it is discarded. This procedure is designed to offer a lower bound of functionality preservation within the feature-space.

\vspace{-3mm}
\section{Black-box Query Attack Results}
\label{sec:blackboxattack}
In the black-box scenario, the attacker can only access the predicted output of the oracle. \hlpink{Hence, we conduct the black-box query attack following the black-box attack strategy in \mbox{Appendix~\ref{appendix:queryattackalgorithm}} and the procedure outlined in \mbox{Section~\ref{sec:expsetup}} for performing query attacks. That is, the black-box attack selects the features to perturb in malware samples in a randomized manner.} We conduct the attack against each \emph{non-stateful} defense as is, to establish the baseline performance of the prediction models. We then evaluate and compare each combination of stateful defense and prediction model, including both MalProtect configurations.

Figure~\ref{figure:BlackBoxFull} shows the evasion rate of the black-box query attack versus \(n_{max}\) \hlpink{(see \mbox{Appendix~\ref{appendix:extendedresults}} for extended results)}. %
Against non-stateful defenses (that is, the prediction models as they are), the black-box query attack can achieve a 100\% evasion rate in some cases, with the average evasion rate sitting at 70+\% across the datasets. These results clearly demonstrate that the non-stateful defenses do not provide adequate protection in this attack scenario, despite being defenses in their own right \hlpink{as they do not have any mechanism to detect attacks in progress. Thus, an attacker can continue to query them and iteratively develop successful adversarial examples.} Out of the prediction models, only veto voting limits the effectiveness of the attack to some extent\hlpink{, likely because of a sensitive constituent model. Against veto voting, the attack achieves} a maximum evasion rate of \(\approx 50\)\% across the datasets. Despite this, around one in two to three queries can still achieve evasion against this model.

\begin{figure*}[!b]
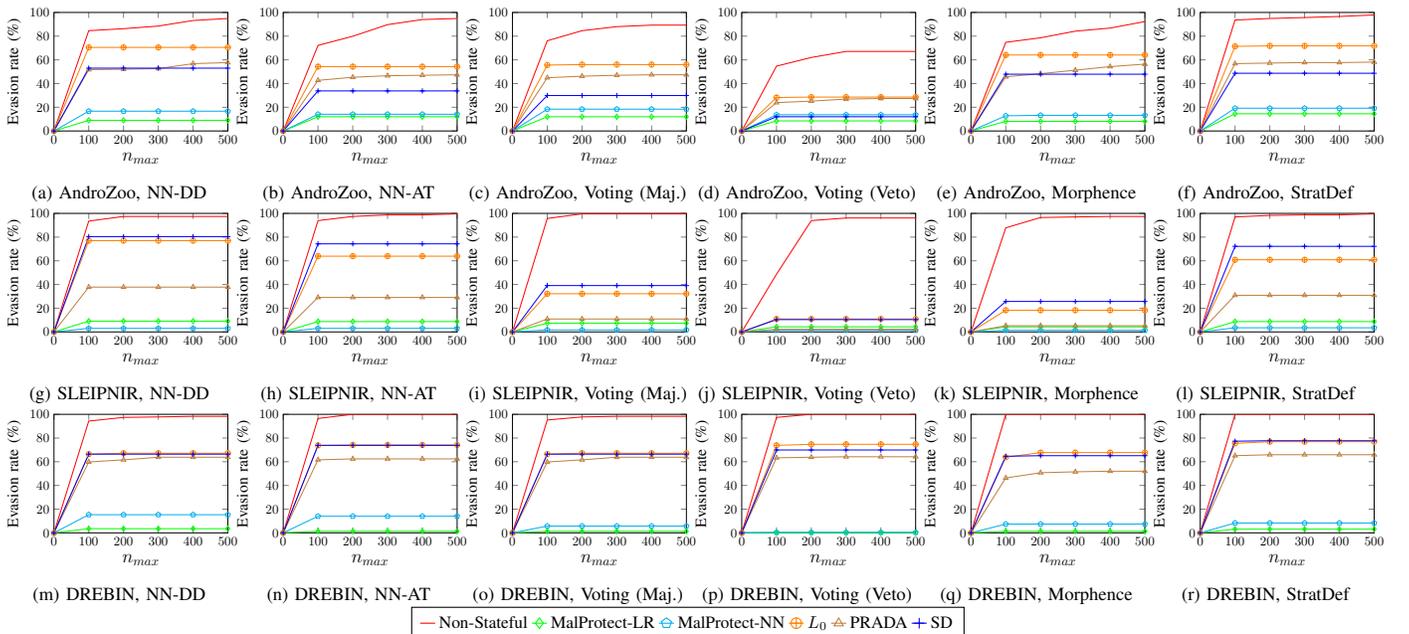

	\vspace{-4mm}
		\centering
		 \advance\leftskip-0.2cm
			
			\begin{subfigure}[b]{0.17\textwidth}

				\SmallerCaption{DREBIN, StratDef}
			\end{subfigure}
			
			\ref{GrayBoxFullLegend}
	\vspace{-2mm}
			\caption{Evasion rate vs. \hlpink{maximum number of queries} (\(n_{max}\)) of gray-box query attack against non-stateful defenses (prediction models) and each combination of prediction model and stateful defense.}
			\label{figure:GrayBoxFull}
			
	\vspace{-4mm}
		\end{figure*}

Meanwhile, our stateful defense, MalProtect, significantly decreases the attack success, with peak reductions in the evasion rates for AndroZoo, SLEIPNIR, and DREBIN of 84\%, 96\% and 98\%, respectively. This highlights the benefits (and necessity) of a stateful defense for this domain, \hlpink{as attacks can be detected using different methods of analysis}. Comparing the two MalProtect configurations, MalProtect-LR and MalProtect-NN exhibit similar performance, with both configurations able to detect attacks after 5, 7, and 2 queries for AndroZoo, SLEIPNIR, and DREBIN, respectively. This level of performance is achieved without greatly compromising other metrics, such as the accuracy or false positive rate on legitimate queries (as we show in Section~\ref{sec:furtherfindings} later). 

Meanwhile, the other stateful defenses that we evaluate are significantly less effective against this attack. For AndroZoo, \(L_{0}\) offers the mildest performance in terms of adversarial robustness, while SD offers slightly improved performance, but only marginally. In the attack's weakest performance against veto voting, we find that SD performs similarly to MalProtect for AndroZoo and SLEIPNIR, but its performance is much inferior in other cases (with other prediction models) where MalProtect demonstrates robustness\hlpink{, likely because SD is unable to detect attacks using techniques beyond similarity detection}. For SLEIPNIR, PRADA offers better performance than the prior stateful defenses, though only marginally in most cases, while \(L_{0}\) and SD seem similar in their defensive robustness. \hlpink{Nonetheless, their performance is much inferior compared to %
MalProtect.} For DREBIN, we observe that the three prior stateful defenses exhibit similar performance in all cases. %
Overall, the black-box query attack can achieve a 60+\% evasion rate in most cases against prior stateful defenses, compared with a peak evasion rate of \(\approx\) 18\% for MalProtect. Hence, MalProtect provides demonstrably better protection against the black-box query attack.

\vspace{-2mm}

\section{Gray-box Query Attack Results}
\label{sec:grayboxattack}
\vspace{-1mm}
Prior work has stressed the importance of evaluating defenses against stronger adversaries \cite{carlini2019evaluating}. Therefore, we next evaluate MalProtect's performance against the gray-box query attack and compare it with other stateful defenses in that attack scenario. 
Under this scenario, the attacker has access to the training data of the prediction models and has more information about the features. The gray-box attack strategy \hlpink{that we use (see \mbox{Appendix~\ref{appendix:queryattackalgorithm}})} therefore utilizes information about the frequency of features in benign samples to determine the order of transplantation, resulting in a more effective attack. %
As before, we conduct the attack against each non-stateful defense to establish a baseline performance, and against each combination of stateful defense and prediction model.

Figure~\ref{figure:GrayBoxFull} shows the evasion rate of the gray-box query attack against all combinations of stateful defenses and prediction models, versus \(n_{max}\) \hlpink{(see \mbox{Appendix~\ref{appendix:extendedresults}} for extended results)}. %
Against the non-stateful defenses, this stronger attack achieves significantly greater evasion across all the datasets, with 100\% evasion rate in many cases. \hlpink{This is because the non-stateful defenses have no mechanism to detect that they are under attack; against a stronger attack, as in this scenario, they exhibit greater vulnerability.} %
\hlpink{Intriguingly,} the greatest difference from the black-box query attack can be observed in the results against veto voting, which was the least evaded non-stateful defense in Section~\ref{sec:blackboxattack}. Whereas the black-box query attack peaked at \(\approx 40\%\) evasion rate against this model (without any stateful defense), the gray-box query attack achieves \(60+\%\) evasion rate.

As far as protection by stateful defenses is concerned, MalProtect performs the best%
. In the best case for the configurations, it reduces the average evasion rate across all models from a peak of 84\% to 13\% for AndroZoo, 94\% to 5\% for SLEIPNIR, and 91\% to 6\% for DREBIN, while only taking \(\approx \) 6, 6, and 2  queries to detect an attack for AndroZoo, SLEIPNIR, and DREBIN, respectively. Like the black-box query attack, MalProtect-LR and MalProtect-NN offer similar performance, with MalProtect-NN slightly outperforming MalProtect-LR in some cases and vice-versa. In fact, the gray-box query attack only achieves a maximum evasion rate of \(\approx\) 20\% against MalProtect across the board. \hlpink{This is higher than the black-box attack results, though expected, since the gray-box attack represents a more hostile threat where the attacker has more knowledge and information.}

Meanwhile, for the other stateful defenses, we observe trends similar to those seen in the results for the black-box query attack. %
For AndroZoo, SD outperforms \(L_{0}\) and PRADA, with average evasion rates of 57.5\%, 46.9\%, and 37.5\% against \(L_{0}\), PRADA, and SD, respectively. %
For SLEIPNIR, PRADA outperforms \(L_{0}\) and SD for the majority of prediction models, a trend that was also observed in the black-box query attack results. Here, the average evasion rates of the attack sit at 43.8\%, 19.3\%, and 50.3\% for \(L_{0}\), PRADA, and SD, respectively. Although PRADA \hlpink{appears to} offer some robustness, it underperforms when other metrics are also considered (\hlpink{especially in false positives,} as we show later in Section~\ref{sec:furtherfindings}). For DREBIN, the prior stateful defenses perform similarly with a more balanced performance. That is, they perform equally poorly, with the gray-box query attack able to achieve average evasion rates of 71.3\%, 62\%, and 69.9\% for \(L_{0}\), PRADA, and SD, respectively, across all configurations. Interestingly, once again, PRADA (minimally) surpasses \(L_{0}\) and SD in terms of robustness, though the poor general performance of these prior stateful defenses seen for both black-box and gray-box results reaffirms that relying on a single similarity or OOD detection mechanism is inadequate for this domain. %
In our domain, attackers use different techniques to generate adversarial examples that often include replacing features rather than making small detectable perturbations.%

Further evaluating the evasion rate versus \(n_{max}\), 100 or fewer queries are enough to achieve attack success in most cases. Query attacks in domains using continuous features (e.g., image recognition) may require substantially more queries \cite{chen2020hopskipjumpattack} to achieve attack success compared with domains that use discrete features. This is because perturbations in a discrete feature-space (e.g., 0 to 1) have a greater effect on the final prediction, requiring fewer of them to accomplish evasion than perturbations made per query  in a continuous feature-space (e.g., +0.01 to some feature). In fact, in Appendix~\ref{appendix:additionalqueryattacks}, we demonstrate the ineffectiveness of query attacks for other domains when applied to ML-based malware detection.

\vspace{-2mm}
\section{Interpretability \& Adaptive Attack}
\label{sec:interp}
\subsection{Interpretability of MalProtect}
As ML is being applied more to the cybersecurity domain, a key challenge is the interpretability of predictions made by models \mbox{\cite{nadeem2022sok}}. As an initial step towards addressing this in stateful defenses, MalProtect produces interpretable decisions. By observing the scores produced by each threat indicator, analysts can better understand how MalProtect made a particular decision, as higher scores for an indicator give clues about what is anomalous about queries. 

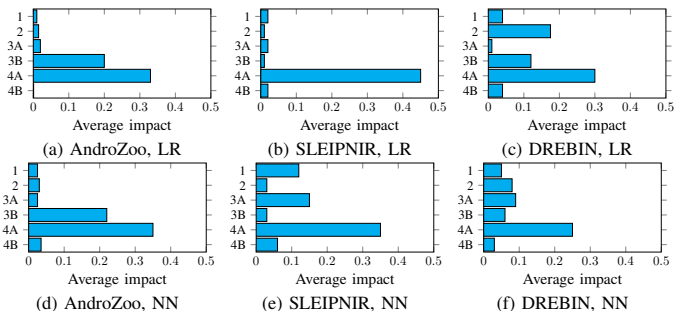
\begin{figure}[!htbp]
    \vspace{-2mm}
    \centering
\advance\leftskip-0.3cm
    \begin{subfigure}[b]{0.16\textwidth}
    \begin{tikzpicture}[scale=0.49]
        \begin{axis}[
            xbar,
            xlabel={Average impact},
            xmin=0,
            xmax=0.5,
            symbolic y coords={4B,4A,3B,3A,2,1},
            ytick=data,
            xtick={0,0.1,0.2,0.3,0.4,0.5},
            nodes near coords align={horizontal},
            width=6.4cm,
            height=4cm,xlabel style = {font=\large}
            ]
        \addplot[fill=cyan] coordinates {(0.01,1)(0.015,2)(0.02,3A)(0.2,3B)(0.33,4A)(0.001,4B)};
        \end{axis}
    \end{tikzpicture}
    \vspace{-6mm}
    \SmallerCaption{AndroZoo, LR}
    \end{subfigure}
    \begin{subfigure}[b]{0.16\textwidth}
    \begin{tikzpicture}[scale=0.49]
        \begin{axis}[
            xbar,
            xlabel={Average impact},
            xmin=0,
            xmax=0.5,
            symbolic y coords={4B,4A,3B,3A,2,1},
            ytick=data,
            xtick={0,0.1,0.2,0.3,0.4,0.5},
            nodes near coords align={horizontal},
            width=6.4cm,
            height=4cm,xlabel style = {font=\large}
            ]
        \addplot[fill=cyan] coordinates {(0.02,1)(0.01,2)(0.02,3A)(0.01,3B)(0.45,4A)(0.02,4B)};
        \end{axis}
    \end{tikzpicture}
    \vspace{-6mm}
    \SmallerCaption{SLEIPNIR, LR}
    \end{subfigure}
    \begin{subfigure}[b]{0.16\textwidth}
    \begin{tikzpicture}[scale=0.49]
        \begin{axis}[
            xbar,
            xlabel={Average impact},
            xmin=0,
            xmax=0.5,
            symbolic y coords={4B,4A,3B,3A,2,1},
            ytick=data,
            xtick={0,0.1,0.2,0.3,0.4,0.5},
            nodes near coords align={horizontal},
            width=6.4cm,
            height=4cm,xlabel style = {font=\large}
            ]
        \addplot[fill=cyan] coordinates {(0.04,1)(0.175,2)(0.01,3A)(0.12,3B)(0.3,4A)(0.04,4B)};
        \end{axis}
    \end{tikzpicture}
    \vspace{-6mm}
    \SmallerCaption{DREBIN, LR}
    \end{subfigure}
    
    \begin{subfigure}[b]{0.16\textwidth}
    \begin{tikzpicture}[scale=0.49]
        \begin{axis}[
            xbar,
            xlabel={Average impact},
            xmin=0,
            xmax=0.5,
            symbolic y coords={4B,4A,3B,3A,2,1},
            ytick=data,
            xtick={0,0.1,0.2,0.3,0.4,0.5},
            nodes near coords align={horizontal},
            width=6.4cm,
            height=4cm,xlabel style = {font=\large}
            ]
        \addplot[fill=cyan] coordinates {(0.025,1)(0.03,2)(0.025,3A)(0.22,3B)(0.35,4A)(0.035,4B)};
        \end{axis}
    \end{tikzpicture}
    \vspace{-6mm}
    \SmallerCaption{AndroZoo, NN}
    \end{subfigure}
    \begin{subfigure}[b]{0.16\textwidth}
    \begin{tikzpicture}[scale=0.49]
        \begin{axis}[
            xbar,
            xlabel={Average impact},
            xmin=0,
            xmax=0.5,
            symbolic y coords={4B,4A,3B,3A,2,1},
            ytick=data,
            xtick={0,0.1,0.2,0.3,0.4,0.5},
            nodes near coords align={horizontal},
            width=6.4cm,
            height=4cm,xlabel style = {font=\large}
            ]
        \addplot[fill=cyan] coordinates {(0.12,1)(0.03,2)(0.15,3A)(0.03,3B)(0.35,4A)(0.06,4B)};
        \end{axis}
    \end{tikzpicture}
    \vspace{-6mm}
    \SmallerCaption{SLEIPNIR, NN}
    \end{subfigure}
    \begin{subfigure}[b]{0.16\textwidth}
    \begin{tikzpicture}[scale=0.49]
        \begin{axis}[
            xbar,
            xlabel={Average impact},
            xmin=0,
            xmax=0.5,
            symbolic y coords={4B,4A,3B,3A,2,1},
            ytick=data,
            xtick={0,0.1,0.2,0.3,0.4,0.5},
            nodes near coords align={horizontal},
            width=6.4cm,
            height=4cm,xlabel style = {font=\large}
            ]
        \addplot[fill=cyan] coordinates {(0.05,1)(0.08,2)(0.09,3A)(0.06,3B)(0.25,4A)(0.03,4B)};
        \end{axis}
    \end{tikzpicture}
    \vspace{-6mm}
    \SmallerCaption{DREBIN, NN}
    \end{subfigure}
    \vspace{-2mm}
    \caption{Influence of MalProtect (LR \& NN) threat indicators. \hlpink{The average impact on model output magnitude is shown for each threat indicator.}}
    \label{figure:featureimportance}
    \vspace{-4mm}
    \end{figure}

The \emph{influence} of each threat indicator on MalProtect's decision can be examined by analyzing the \emph{global feature importance} in MalProtect's decision model. Recall that each threat indicator is a feature of the decision model. %
Therefore, we can assess the global feature importance using SHAP \cite{lundberg2017unified}, which is a widely-used framework for interpreting ML predictions. Each feature of the decision model of the MalProtect configurations is given an importance value by SHAP at a global level, which produces the data observable in Figure~\ref{figure:featureimportance}.

Figure~\ref{figure:featureimportance} shows that the Indicator 4A has the greatest influence on MalProtect's predicted output across all decision models and datasets, which indicates that a significantly high autoencoder loss is more likely to affect the predicted output. Meanwhile, the importance of the other indicators is more balanced. Intriguingly, the indicators based on similarity detection seem less influential, which is interesting because other stateful defenses rely solely on similarity detection to detect attacks. In malware detection, not only do we show that attackers can evade stateful defenses relying solely on such detection methods, but we further show that MalProtect rightly considers those indicators as less important. %

Intriguingly, a capable attacker could leverage this information to evade MalProtect. In an adaptive attack (i.e., white-box attack) scenario \cite{carlini2019evaluating,tramer2020adaptive}, the attacker has complete knowledge of MalProtect and, thus, knowledge of how much each indicator contributes to the final decision. With this information, an attacker could craft an adversarial example that evades both MalProtect and the underlying prediction model. We next examine how such an attack could be performed.

\noindent{\textbf{Attack Rationale.}} In the adaptive attacker scenario, the attacker knows precisely how MalProtect operates, its internal workings, and other pertinent information related to the threat indicators. At a high level, to evade MalProtect's detection system, adversarial queries must appear as legitimate as possible and convince the indicators that the queries are distinct yet \emph{normal}. Additionally, adversarial queries must remain within the distribution of legitimate queries. For example, we know that the Indicator 4A is the most influential in the final prediction, and therefore it would be essential to ensure that the autoencoder loss is not abnormally high. In essence, queries must remain far apart from each other while retaining their original malicious functionality and appearing legitimate. Importantly, any generated adversarial example must also evade the underlying prediction model; otherwise, such an attack would be of no use. Attackers must bypass all components of the target model to be successful \cite{apruzzese2023realgradients}. %

\noindent{\textbf{Adaptive Attack Strategy.}} We modify the gray-box attack strategy to produce the adaptive attack strategy (see Appendix~\ref{appendix:queryattackalgorithm}). Firstly, we limit the number of features that can be perturbed in a single iteration of the attack so as not to exhaust possible perturbations early on. This frees up perturbations to be used in later iterations of the attack if earlier queries cannot achieve evasion. While this may increase the query distance and make queries appear less anomalous, other indicators (e.g., Indicator 4A) may still be able to trigger attack detection. To deal with this, the adaptive attack strategy also \emph{removes} a proportion (\(p\)) of features at each iteration. This means that queries will be more distant, with fewer shared and enabled features, while conforming to the distribution of legitimate queries and the training data. For example, a \emph{combination of features} that may cause an increase in the autoencoder loss (such that it appears anomalous) may be removed. Only the AndroZoo and DREBIN datasets support the removal of features while preserving functionality within the feature-space. Therefore, the adaptive attack strategy can only be applied to these datasets.

\begin{figure}[!htbp]
	\vspace{-3.5mm}
	\centering
	\begin{subfigure}[b]{0.25\textwidth}
	\begin{tikzpicture}[scale=0.49]
		\begin{axis}[
			xlabel={\% Features removed (\(p\))},
			ylabel={Average evasion rate (\%)}, 
			xmin=0, xmax=100,
			ymin=0, ymax=100,
			xtick={0,20,40,60,80,100},
			ymajorgrids=false,
			legend pos=south west,
			legend style={nodes={scale=0.65, transform shape}},
			cycle list name=color list,
			legend to name=interpleg,
			legend columns=6,
			height=4.7cm,
			width=8cm,xlabel style = {font=\large},ylabel style = {font=\large}
		]
		
		\addplot[color=green,mark=diamond]
			coordinates {
		(0,11.7)(10,14.5)(20,14.5)(30,14.1)(40,13.2)(50,15.4)(60,15.8)(70,12.8)(80,16.7)(90,15.4)(100,15)
			};
		\addlegendentry{MalProtect-LR}

		\addplot[color=cyan,mark=pentagon]
			coordinates {
	(0,9.9)(10,12)(20,12)(30,12.4)(40,9)(50,12)(60,11.1)(70,11.1)(80,9.4)(90,12.8)(100,12.8)
			};
		\addlegendentry{MalProtect-NN}

		\addplot[color=orange,mark=oplus]
			coordinates {
			(0,54.3)(10,48.5)(20,48.6)(30,49.8)(40,49.6)(50,48.7)(60,49.9)(70,46.4)(80,44.2)(90,46.1)(100,46.3)
			};
		\addlegendentry{$L_{0}$}
		
		\addplot[color=brown,mark=triangle]
			coordinates {
		(0,47.4)(10,36.3)(20,32.1)(30,36)(40,32.8)(50,31.2)(60,32.3)(70,29.8)(80,30.8)(90,31.3)(100,31.3)
			};
		\addlegendentry{PRADA}

		\addplot[color=blue,mark=+]
			coordinates {
		(0,33.7)(10,30.9)(20,31.9)(30,33.4)(40,33)(50,32.5)(60,32.2)(70,31.1)(80,31.7)(90,31.9)(100,31.9)
			};
		\addlegendentry{SD}
		
		\end{axis}
	\end{tikzpicture}
	\vspace{-2mm}
	\caption{AndroZoo}
	\end{subfigure}
	\hskip -1ex
	\begin{subfigure}[b]{0.22\textwidth}
	\begin{tikzpicture}[scale=0.49]
		\begin{axis}[
			xlabel={\% Features removed (\(p\))},
			ylabel={Average evasion rate (\%)}, 
			xmin=0, xmax=100,
			ymin=0, ymax=100,
			xtick={0,20,40,60,80,100},
			ymajorgrids=false,
			legend pos=south west,
			legend style={nodes={scale=0.65, transform shape}},
			cycle list name=color list,
			legend to name=interpleg,
			legend columns=6,
			height=4.7cm,
			width=8cm,xlabel style = {font=\large},ylabel style = {font=\large}
		]
		
		\addplot[color=green,mark=diamond]
			coordinates {
			(0,0.7)(10,2.2)(20,1.7)(30,1.3)(40,1.7)(50,1.3)(60,1.3)(70,1.3)(80,1.3)(90,1.3)(100,1.3)
			};
		\addlegendentry{MalProtect-LR}

		\addplot[color=cyan,mark=pentagon]
			coordinates {
		(0,13.4)(10,9.2)(20,9.6)(30,9.6)(40,9.2)(50,10)(60,11.4)(70,9.2)(80,9.6)(90,10)(100,8.3)
			};
		\addlegendentry{MalProtect-NN}

		\addplot[color=orange,mark=oplus]
			coordinates {
			(0, 69.578)(10, 78.093)(20, 75.328)(30, 74.454)(40, 73.435)(50, 72.416)(60, 72.707)(70, 71.761)(80, 72.635)(90, 70.815)(100, 70.524)
			};
		\addlegendentry{$L_{0}$}
		
		\addplot[color=brown,mark=triangle]
			coordinates {
			(0, 58.006)(10, 63.246)(20, 61.426)(30, 61.863)(40, 62.154)(50, 62.154)(60, 60.189)(70, 63.028)(80, 61.354)(90, 63.1)(100, 64.265)
			};
		\addlegendentry{PRADA}
		\addplot[color=blue,mark=+]
			coordinates {
			(0, 69.578)(10, 76.201)(20, 71.252)(30, 69.36)(40, 68.632)(50, 66.739)(60, 67.031)(70, 63.464)(80, 64.41)(90, 63.246)(100, 64.702)
			};
		\addlegendentry{SD}
		\end{axis}
	\end{tikzpicture}
	\vspace{-6mm}
	\caption{DREBIN}
	\end{subfigure}
	\ref{interpleg}
	\vspace{-2mm}
	\caption{Average evasion rate vs. \(p\) for adaptive attack. The average is across the maximum number of queries permitted.} %
	\label{figure:interp}
	\vspace{-3mm}
	\end{figure}
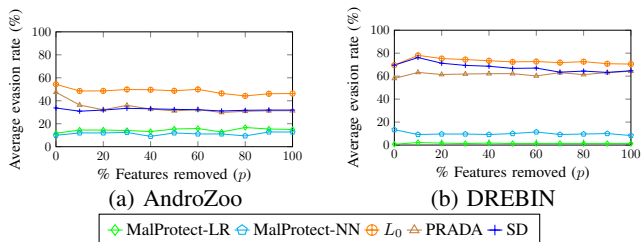

\noindent{\textbf{Results.}} Figure~\ref{figure:interp} shows the evasion rate of the adaptive attack against the stateful defenses (averaged across the maximum number of queries permitted, up to 500) versus the percentage of features removed (\(p\)). Recall that successful adversarial examples evade each stateful defense as well as the underlying prediction model, which is NN-AT.

For both datasets, the adaptive attack fails to achieve significant increases in the evasion rate against either MalProtect configuration. In fact, in some cases, the average evasion rate decreases as \(p\) increases. This is likely because the (benign) features selected to cross the decision boundary are in fact removed. %
This attack causes other stateful defenses to exhibit similar performance to the gray-box query attack, with an average evasion rate of 72.9\%, 61.9\%, and  67.7\% for \(L_{0}\), PRADA, and SD, respectively, and consistent evasion rate across different values of \(p\). This is expected, as the adaptive attack is designed to target MalProtect and not the other stateful defenses (which are included as a comparison here). 

\vspace{-2mm}
\section{Beyond Adversarial Robustness}
\label{sec:furtherfindings}
We next evaluate the stateful defenses under different system conditions using metrics beyond the evasion rate. To do this, we query stateful defenses with benign and malicious queries, some of which are adversarial. This is imperative, as the original task of classifying queries as benign or malicious must also be done properly.

\noindent{\textbf{Procedure.}} %
We examine the performance of stateful defenses in different system conditions with different levels of adversity. For this, we generate adversarial examples using the transferability attack strategy described in \mbox{Section~\ref{sec:expsetup}}. Each stateful defense (with NN-AT as the prediction model) is then queried 1000+ times with values of $0.1 \leq k \leq 0.9$, where $k$ represents the adversarial intensity. For example, at $k=0.1$ --- which represents a less adversarial environment --- 10\% of all queries are adversarial examples, while the remaining 90\% are equally split between benign and \emph{non-adversarial} malware samples from the test set. As $k$ increases, the system conditions become more hostile as the system faces more adversarial queries. We do not consider $k=0$ (no adversarial examples) or $k=1$ (only adversarial examples), as such environments are less likely. As before, the query history is initialized as necessary with random samples from the training data.

\begin{figure}[!htbp]
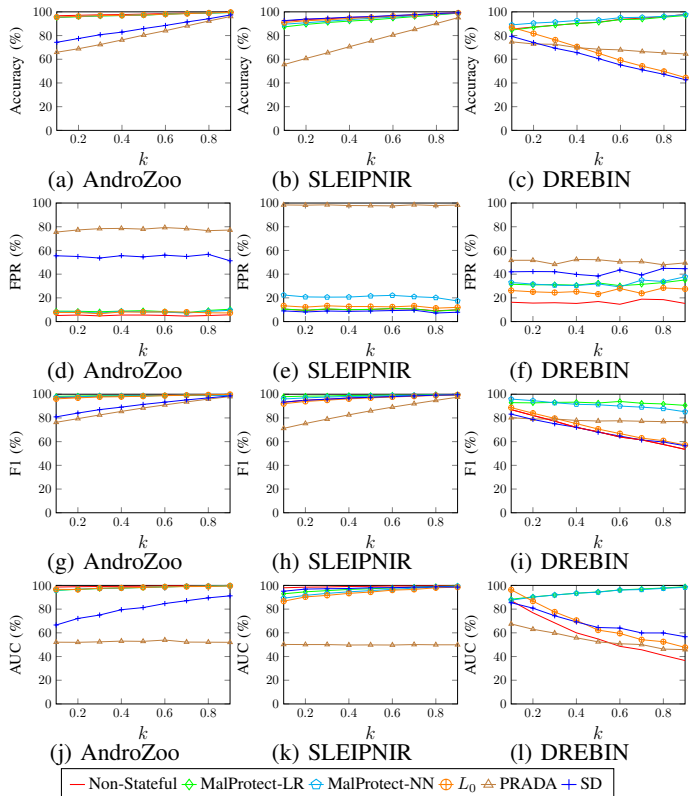

	\vspace{-3.5mm}
		\centering
			 \advance\leftskip-0.5cm
		\begin{subfigure}[b]{0.16\textwidth}

			\vspace{-6.5mm}
			\caption{AndroZoo}
		\end{subfigure}
		\begin{subfigure}[b]{0.16\textwidth}
			%
			\vspace{-6.5mm}
			\caption{SLEIPNIR}
		\end{subfigure}
		\begin{subfigure}[b]{0.16\textwidth}
			%
			\vspace{-6.5mm}
			\caption{DREBIN}
		\end{subfigure}
	
		\begin{subfigure}[b]{0.16\textwidth}
			%
			\vspace{-6.5mm}
			\caption{AndroZoo}
		\end{subfigure}
		\begin{subfigure}[b]{0.16\textwidth}
			%
			\vspace{-6.5mm}
			\caption{SLEIPNIR}
		\end{subfigure}
		\begin{subfigure}[b]{0.16\textwidth}
			%
			\vspace{-6.5mm}
			\caption{DREBIN}
		\end{subfigure}
		
			\begin{subfigure}[b]{0.16\textwidth}
			%
			\vspace{-6.5mm}
			\caption{AndroZoo}
		\end{subfigure}
		\begin{subfigure}[b]{0.16\textwidth}
			%
			\vspace{-6.5mm}
			\caption{SLEIPNIR}
		\end{subfigure}
		\begin{subfigure}[b]{0.16\textwidth}
			%
			\vspace{-6.5mm}
			\caption{DREBIN}
		\end{subfigure}
		
		\begin{subfigure}[b]{0.16\textwidth}
			%
			\vspace{-6.5mm}
			\caption{DREBIN}
		\end{subfigure}
		\ref{grayboxfullfprglegend}
	\vspace{-2mm}
		 \caption{Accuracy, FPR, F1 \& AUC vs. \(k\) per  stateful defense.}%
		\label{figure:grayboxfullfprg}
	\vspace{-3.5mm}
	\end{figure}

\noindent{\textbf{Results.}} \mbox{Figure~\ref{figure:grayboxfullfprg}} shows the performance of each stateful defense versus $k$. In particular, MalProtect exhibits reasonably stable performance across the datasets considering all metrics, with the accuracy and F1 remaining greater than 90\%, demonstrating MalProtect's ability to properly classify user queries. %
Intriguingly, other stateful defenses also appear to provide decent performance in terms of these metrics in some cases. For AndroZoo and SLEIPNIR, $L_{0}$, PRADA, and SD all exhibit relatively stable accuracy and F1 as $k$ increases. This is possibly because the stateful defenses have it easier to detect the (less evasive) adversarial examples, which constitute most of the queries, as they are generated by attacks not for this domain. Recall that the perturbations applied by the attacks to generate the adversarial examples may be reversed if features are modified such that functionality is affected; this would limit attack success. Regardless, these stateful defenses do not perform as well as MalProtect. For DREBIN, we observe the decline of the non-stateful NN-AT model and stateful defenses except MalProtect when these metrics are considered.

Importantly, though, additional metrics must be examined to obtain a comprehensive assessment of performance. Analysis of the AUC reveals MalProtect's superiority in distinguishing the benign and malware classes across the datasets without disrupting the classification of legitimate queries. Meanwhile, PRADA and SD exhibit significantly lower AUC in some cases (especially as $k$ increases), implying that their decisions are no better than flipping a coin, which is contradictory to the higher accuracy and F1 that they exhibit. This is possibly because as $k$ increases, the number of malware queries also increases in the test set because of more adversarial queries. If PRADA, for example, decides that most queries are adversarial (and therefore malware), this would lead to high accuracy, even if it misclassifies benign queries. Therefore, PRADA seems less capable of separating between the classes in this domain. This is further evidenced through an examination of the false positive rate (FPR) next.

In the malware detection domain, as we have explained before, a low FPR is essential to ensure a reliable service for users \mbox{\cite{grosse2017adversarial, stokes2017attack, yang2017malware,10.1145/3484491}}. %
Although a stateful defense may improve protection against attacks, the FPR may also increase compared with the underlying prediction model. This can be attributed to the sensitivity of detection mechanisms; that is, benign queries may be more frequently misclassified as malicious. Importantly, though, MalProtect offers a low FPR that is as close as possible to the non-stateful prediction model for all datasets as it uses several indicators to predict an attack. Meanwhile, $L_{0}$ is the only other stateful defense that offers comparable performance to MalProtect in this regard. However, as we have seen, it does not work well at defending against query attacks in this domain. Conversely, PRADA exhibits a significantly high FPR for nearly all datasets, which supports the notion that it is merely classifying most queries as malware or adversarial rather than actually distinguishing queries between the classes in this experiment. This is somewhat resembled by SD, with its high FPR for the Android datasets. Overall, across the metrics and datasets, we showcase MalProtect's ability to perform well across a range of different queries.

\vspace{-2mm}

\section{Efficiency of MalProtect vs. Other Defenses}
\label{sec:overhead}
\hlpink{Another important 
aspect is to evaluate MalProtect's  storage and time costs and compare them with those of other defenses.}

\noindent{\textbf{Storage.}} The storage cost scales linearly with the size of the query history, \(Q\)\hlpink{, where queries are stored. The storage costs are the same for any stateful defense that uses a query history of the same size.} In our evaluation, the size of \(Q\) is capped at 10,000, which only consumes \(\approx\) 0.08MB storage for all the datasets. \hlpink{If $|Q|$ is increased to 50K, only $\approx$ 0.4MB would be required for all datasets}, which is %
negligible.

\begin{figure}[!htbp]
\vspace{-3.5mm}
	\centering
	\begin{subfigure}[b]{0.15\textwidth}
		\begin{tikzpicture}[scale=0.49]
			\begin{axis}[
					xlabel={\(|Q|\)},
					ylabel={Time (secs)}, 
					xmin=0, xmax=50000,
					ymin=0, ymax=4,
					xtick={0,10000,20000,30000,40000,50000},
                    xticklabels={0,10K,20K,30K,40K,50K},
					ytick={0,1,2,3,4},
					ymajorgrids=false,
					legend pos=south west,
					legend style={nodes={scale=0.65, transform shape}},
					cycle list name=color list,
					legend to name=worstcaseruntimelegend,
					legend columns=8,
					height=4.7cm,
					width=6.2cm,
					scaled x ticks=false, xlabel style = {font=\Large},ylabel style = {font=\large}
				]
	 
	\addplot[color=green,mark=diamond]
		coordinates {
		(0,0)(10000,0.353)(20000,0.707)(30000,1.06)(40000,1.414)(50000,1.767)
		};
	\addlegendentry{MalProtect}
	
	\addplot[color=orange,mark=oplus]
		coordinates {
	(0,0)(10000,0.192)(20000,0.385)(30000,0.577)(40000,0.769)(50000,0.962)
		};
	\addlegendentry{$L_{0}$}
	
	\addplot[color=brown,mark=triangle]
		coordinates {
			(0,0)(10000,0.031)(20000,0.061)(30000,0.092)(40000,0.123)(50000,0.154)
		};
	\addlegendentry{PRADA}		 
					 
	\addplot[color=blue,mark=+]
		coordinates {
		(0,0)(10000,0.248)(20000,0.495)(30000,0.743)(40000,0.991)(50000,1.239)
		};
	\addlegendentry{SD}
	
			\end{axis}
		\end{tikzpicture}
		\caption{AndroZoo}
	\end{subfigure}
	\begin{subfigure}[b]{0.15\textwidth}
		\begin{tikzpicture}[scale=0.49]
		    \begin{axis}[
					xlabel={\(|Q|\)},
					ylabel={Time (secs)}, 
					xmin=0, xmax=50000,
					ymin=0, ymax=4,
					xtick={0,10000,20000,30000,40000,50000},
                    xticklabels={0,10K,20K,30K,40K,50K},
					ytick={0,1,2,3,4},
					ymajorgrids=false,
					legend pos=south west,
					legend style={nodes={scale=0.65, transform shape}},
					cycle list name=color list,
					legend to name=worstcaseruntimelegend,
					legend columns=8,
					height=4.7cm,
					width=6.2cm,
					scaled x ticks=false, xlabel style = {font=\Large},ylabel style = {font=\large}
				]
	 
	\addplot[color=green,mark=diamond]
		coordinates {
		(0,0)(10000,0.351)(20000,0.702)(30000,1.054)(40000,1.405)(50000,1.756)
		};
	\addlegendentry{MalProtect}

	\addplot[color=orange,mark=oplus]
		coordinates {
		(0,0)(10000,0.214)(20000,0.428)(30000,0.641)(40000,0.855)(50000,1.069)
		};
	\addlegendentry{$L_{0}$}
	
	\addplot[color=brown,mark=triangle]
		coordinates {
		(0,0)(10000,0.023)(20000,0.047)(30000,0.07)(40000,0.093)(50000,0.116)
		};
	\addlegendentry{PRADA}		 
					 
	\addplot[color=blue,mark=+]
		coordinates {
    	(0,0)(10000,0.272)(20000,0.543)(30000,0.815)(40000,1.087)(50000,1.358)
		};
	\addlegendentry{SD}
	
			\end{axis}
		\end{tikzpicture}
		\caption{SLEIPNIR}
	\end{subfigure}
	\begin{subfigure}[b]{0.15\textwidth}
		\begin{tikzpicture}[scale=0.49]
			\begin{axis}[
					xlabel={\(|Q|\)},
					ylabel={Time (secs)}, 
					xmin=0, xmax=50000,
					ymin=0, ymax=4,
					xtick={0,10000,20000,30000,40000,50000},
                    xticklabels={0,10K,20K,30K,40K,50K},
					ytick={0,1,2,3,4},
					ymajorgrids=false,
					legend pos=south west,
					legend style={nodes={scale=0.65, transform shape}},
					cycle list name=color list,
					legend to name=worstcaseruntimelegend,
					legend columns=8,
					height=4.7cm,
					width=6.2cm,
					scaled x ticks=false, xlabel style = {font=\Large},ylabel style = {font=\large}
				]
	 
	\addplot[color=green,mark=diamond]
		coordinates {
		(0,0)(10000,0.569)(20000,1.137)(30000,1.706)(40000,2.275)(50000,2.844)
		};
	\addlegendentry{MalProtect}
	
	\addplot[color=orange,mark=oplus]
		coordinates {
	    (0,0)(10000,0.623)(20000,1.245)(30000,1.868)(40000,2.49)(50000,3.113)
		};
	\addlegendentry{$L_{0}$}
	
	\addplot[color=brown,mark=triangle]
		coordinates {
		(0,0)(10000,0.095)(20000,0.191)(30000,0.286)(40000,0.381)(50000,0.476)
		};
	\addlegendentry{PRADA}		 
					 
	\addplot[color=blue,mark=+]
		coordinates {
    	(0,0)(10000,0.389)(20000,0.779)(30000,1.168)(40000,1.557)(50000,1.946)
		};
	\addlegendentry{SD}
	
			\end{axis}
		\end{tikzpicture}
		\caption{DREBIN}
	\end{subfigure}

	\ref{worstcaseruntimelegend}
\vspace{-2mm}
	\caption{\hlpink{Worst-case prediction time vs. size of $Q$ ($|Q|$). Importantly, with optimization and distributed computing techniques, the time costs can be drastically reduced.}}%
	\label{figure:worstcaseruntime}
\vspace{-2mm}
\end{figure}
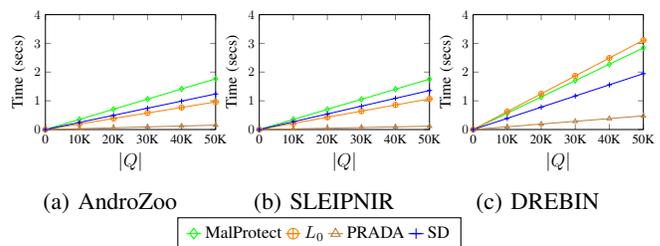

\noindent{\textbf{Prediction Time.}} \hlpink{The \emph{prediction time} refers to the total duration from when a query is received by the system until a prediction is returned to the user. For most non-stateful defenses and ML classifiers, this is nominal. However, there is latency associated with the analysis of the query history by stateful defenses, resulting in longer prediction times. In the worst-case, stateful defenses must analyze the entire query history, whose time cost is the \emph{worst-case prediction time}. \mbox{Figure~\ref{figure:worstcaseruntime}} shows how the worst-case prediction time is affected by $|Q|$.}

\hlpink{%
\mbox{Figure~\ref{figure:worstcaseruntime}} shows that the worst-case prediction-time scales linearly with $|Q|$ for all defenses. %
In our experiments where $|Q|$ is 10,000, the worst-case prediction time for MalProtect sits at $\approx$ 0.4 seconds for AndroZoo and SLEIPNIR, and $\approx$ 0.6 seconds for DREBIN. Both MalProtect configurations have the same overall cost as they have an identical analysis stage, while predictions from the decision model are near-instant. Meanwhile, other stateful defenses such as $L_{0}$, PRADA, and SD have slightly lower worst-case prediction times ($\approx$ 0.03-0.6 seconds across the datasets) for this size of $Q$. In all cases, however, the worst-case prediction times increase linearly with $|Q|$. %
This means that %
with MalProtect, a significant improvement in robustness can be achieved (up to 80\% reduction in evasion rate as shown in our experimental evaluation) at a slightly higher but still linear cost.}

\noindent{\textbf{Optimization.}} 
\hlpink{MalProtect's performance can be easily optimized.  %
For example, by running MalProtect on an NVIDIA A100 GPU, the worst-case prediction time is reduced by over half ($\approx$ 0.21-0.41 seconds) across the datasets when $|Q|=10,000$.} %

\vspace{-2mm}

\section{Conclusion}
\label{sec:conclusion}
In this paper, we presented MalProtect, which is the first stateful defense for adversarial query attacks in the ML-based malware detection domain. As we have shown, ML prediction models and defenses exhibit significant vulnerability to query attacks in this domain. Prior stateful defenses that have been applied to other domains provide little protection here either. Meanwhile, our defense, MalProtect, does not rely solely on a single form of similarity or out-of-distribution detection, as these prior stateful defenses do. Instead, we use several threat indicators and a decision model to detect attacks more effectively. Our evaluation has shown that MalProtect performs well against attacks under various scenarios and also offers more reliable predictions for non-adversarial queries than prior stateful defenses. Furthermore, MalProtect displays resilience even against adaptive attackers. %

In future work, \hlpink{we aim to explore a number of key aspects. Firstly, we aim to better understand the relationship between the size of the query history, the defender's available resources, and the detection effectiveness. Since the query history cannot be infinite, it may be possible for attackers to submit multiple queries in order to move the sliding window to a particular point. An attacker may then dispatch adversarial queries at specific intervals so as to evade detection in a \emph{spaced-out attack}. However, this attack may also be considered less practical for a number of reasons. The attacker would somehow need to know the size of the query history and then spread adversarial queries accordingly, notwithstanding the associated cost of doing so (e.g., time). Moreover, MalProtect is not a defense that is to be deployed in complete isolation. Hence, there is no guarantee that if a spaced-out attack were to succeed in evading the sliding window for detection, the underlying prediction model would be evaded.}

\hlpink{Secondly, we seek to understand} how additional indicators could be added to increase protection, such as cyber-threat intelligence \cite{shu2018threat,zhu2018chainsmith}. Moreover, there is an open research direction regarding \emph{concept drift}. This relates to the constant evolution of malware, which makes it difficult to detect unseen behavior \cite{203684, 10.1145/3183440.3195004, 8802672}, leading to unsustainable models. Prior work has suggested retraining a model regularly \mbox{\cite{10.1007/978-3-642-04342-0_2, robotics8030050}}. In the case of MalProtect, the defender may need to regularly evaluate the indicators \hlpink{and the decision model for predicting attacks}. One potential way to do this would be to couple MalProtect with a detection framework for detecting when such modifications are necessary \cite{203684}. \hlpink{A further intriguing research direction is exploring the effectiveness of \emph{online learning} in stateful defenses. With such a mechanism, the underlying prediction models could be updated on queries that are considered adversarial by the detection system in real-time.}

\vspace{-2mm}
\bibliographystyle{IEEEtran}
\bibliography{bib.bib}

\begin{appendices}

\vspace{-3mm}
\section{MalProtect}
\label{appendix:configurations}
\vspace{-3mm}

\begin{table}[H]
\centering
\scalebox{0.8}{
\begin{tabular}{ll} 
\hline
Model & Parameters \\
\hline 
\makecell[l]{Logistic Regression\\(MalProtect-LR)} & \makecell[l]{No configurable parameters.} \\
\hdashline
\makecell[l]{Neural Network\\(MalProtect-NN)} & \makecell[l]{4 fully-connected layers (128 (Relu),  64 (Relu),\\ 32 (Relu), 2 (Softmax))} \\
\hline
\end{tabular}
}
\caption{Decision models for MalProtect configurations.}
\end{table}
\vspace{-3mm}

\vspace{-2mm}

\section{Prediction Models}
\label{appendix:modelarchs}
We use six non-stateful defenses as the prediction models. These are evaluated without any stateful protection and then evaluated in combination with each stateful defense.

\begin{table}[!htbp]
\centering
\scalebox{0.8}{
\begin{tabular}{ll} 
\hline
Defense & Configuration/parameters/setup \\
\hline 
Defensive distillation \cite{papernot2016distillation} & \makecell[l]{Neural Network (128 (Relu), 64 (Relu), 32 (Relu),\\ 2 (Softmax) with defensive distillation applied.} \\
\hdashline 
\makecell[l]{Ensemble adversarial\\ training \cite{tramer2017ensemble, rashid2022stratdef}\\ (NN-AT)}  & \makecell[l]{Neural Network (128 (Relu), 64 (Relu), 32 (Relu),\\ 2 (Softmax)). Adversarially-trained up to 25\% size of\\ training data.} \\
\hdashline
Morphence \cite{amich2021morphence} & \makecell[l]{\(n=4\), \(p=3\), \(Q_{max}=1000\).} \\
\hdashline
StratDef \cite{rashid2022stratdef} & \makecell[l]{Variety-GT using same models as original paper.\\ Assumed strong attacker and \(\alpha=1\).} \\
\hdashline
Voting (Majority \& Veto) & \makecell[l]{Using same models as StratDef. } \\
\hline
\end{tabular}
}
\caption{Architectures of prediction models.}
\vspace{-3.5mm}
\end{table}

\vspace{-5mm}

\section{Vanilla Models}
\label{appendix:vanilamodels}
The following vanilla models are used in some instances (e.g., to generate adversarial examples, see Section~\ref{sec:expsetup}).
\begin{table}[!htbp]
\centering
\scalebox{0.8}{
\begin{tabular}{ll} 
\hline
Model & Parameters \\
\hline 
Decision Tree &  max\_depth=5, min\_samples\_leaf=1 \\
\hdashline
Neural Network & \makecell[l]{4 fully-connected layers (128 (Relu), \\  64 (Relu), 32 (Relu), 2 (Softmax))} \\
\hdashline
Random Forest &  max\_depth=100 \\ 
\hdashline
Support Vector Machine &  LinearSVC with probability enabled \\ 
\hline
\end{tabular}
}
\caption{Architectures of vanilla models.}
\end{table}

\vspace{-5mm}

\section{Other Stateful Defenses}
\label{appendix:otherstatefuldefenses}

\noindent{\textbf{\(L_{0}\) defense.}} The \(L_{0}\) defense measures the similarity between feature vectors using the \(L_{0}\) distance. This is akin to similarity detection schemes that use \(L_{p}\) norms in other domains (e.g., \cite{li2020blacklight}). As binary feature vectors are used in the malware detection domain, $L_{0}$ is the most appropriate measure of distance between two such feature vectors, \(X\) and \(X'\) \cite{carlini2017towards, pods}. It measures the number of instances such that \(X_{i} \neq X'_{i}\). For the \(L_{0}\) defense, an attack is detected if there are queries with an \(L_{0}\) distance of less than 10 as lower thresholds may easily miss attacks \cite{li2020blacklight, juuti2019prada, chen2020stateful}.

\noindent{\textbf{PRADA \cite{juuti2019prada}}.} For PRADA, we use \(\delta = 0.9\) for the threshold, as per the original paper \cite{juuti2019prada}.

\noindent{\textbf{Stateful Detection (SD) \cite{chen2020stateful}}.} For SD, the threshold for detecting an attack is derived by calculating the k-neighbor distance for the 0.1 percentile of the training set. At prediction-time, if the mean distance of the k-nearest queries falls below the calculated threshold, an attack is detected. For this, we use \(k=50\) (as per the original paper). The calculated thresholds for detection are as follows:

\begin{table}[!htbp]
\vspace{-2mm}
\centering
\scalebox{0.9}{
\begin{tabular}{lll} 
\hline
AndroZoo & SLEIPNIR & DREBIN \\
\hline 
94.64749019607844 & 44.350758426966294 & 34.37145577151924 \\
\hline
\end{tabular}
}
\caption{Thresholds for detection for SD as per calculations based on original paper.}
\vspace{-5mm}
\end{table}

\vspace{-3mm}

\section{Permitted Perturbations for Android Datasets}
\label{appendix:drebinpermitted}
The AndroZoo \mbox{\cite{Allix:2016:ACM:2901739.2903508}} and DREBIN \mbox{\cite{arp2014drebin}} datasets are based on the Android platform. Both datasets can be divided into eight feature families comprised of extracted static features such as permissions, API calls, hardware requests, and URL requests. Based on their family, features may be addable or removable during attacks to traverse the decision boundary, according to prior work and industry documentation (e.g., \mbox{\cite{li2021framework, 8171381, li2020enhancing, al2018adversarial, pierazzi2020problemspace, rashid2022stratdef}}). However, it is imperative to preserve malicious functionality in the feature-space as a core constraint in this domain. For example, attacks cannot remove features from the manifest file nor intent filter, and component names must be consistently named. Therefore, \mbox{Table~\ref{table:permittedperturbations}} summarizes the permitted perturbations for each feature family for these datasets. This is used to determine whether the perturbations performed by an attack are valid. For example, if a feature belonging to the S1 family is removed by an attack, its original value is restored as it is not permitted to be removed (see \mbox{Section~\ref{sec:expsetup}}).

\begin{table}[H]
\centering
\scalebox{0.9}{
\begin{tabular}{llcc}
    \hline
     & Feature families & \multicolumn{1}{l}{Addition} & \multicolumn{1}{l}{Removal} \\ \hline
    \multirow{4}{*}{manifest} & S1 Hardware & \cmark & \xmark \\
     & S2 Requested permissions & \cmark & \xmark \\
     & S3 Application components & \cmark & \cmark \\
     & S4 Intents & \cmark & \xmark \\ \hline
    \multirow{4}{*}{dexcode} & S5 Restricted API Calls & \cmark & \cmark \\
     & S6 Used permission & \xmark & \xmark \\
     & S7 Suspicious API calls & \cmark & \cmark \\
     & S8 Network addresses & \cmark & \cmark \\ \hline
    \end{tabular}
}
\caption{Permitted perturbations for Android datasets.}
\label{table:permittedperturbations}
\end{table}

\section{Query Attack Strategies}
\label{appendix:queryattackalgorithm}

We use variations of attack strategies from prior work in our domain %
\cite{rosenberg2020query}. These are based on a software transplantation approach where an attacker makes perturbations based on benign samples. We modify prior attack strategies by transplanting multiple features per iteration (rather than one perturbation per iteration). The differences between the black-box and gray-box strategies lie in how perturbations are applied. The black-box attack selects the features to perturb in a randomized manner, while the gray-box attack perturbs features based on their frequency in benign samples in a heuristically-driven approach. \hlpink{Recall that the black-box query attack strategy is used in \mbox{Section~\ref{sec:blackboxattack}} and the gray-box strategy in \mbox{Section~\ref{sec:grayboxattack}}.}

For the adaptive attack against MalProtect \hlpink{(in \mbox{Section~\ref{sec:interp})}}, the number of features that are added in a single perturbation is capped, and from the \emph{removable features} of the query, \(p\)\% of features are removed. This is to make queries as distinct from each other as possible while remaining within the general distribution of other queries.

\vspace{-3mm}

\begin{figure}[!htbp]
  \centering
  \resizebox{0.75\linewidth}{!}{%
    \begin{minipage}{\linewidth}
\begin{algorithm}[H]
\caption{Black-box query attack: for oracle \(O\), malware sample \(X\), set of (randomly-ordered) benign features \(F\), maximum permitted queries \(n_{max}\).}
\label{alg:statisticalqbased}
\textbf{Input:} \(O\), \(X\), \(F\), \(n_{max}\)
\begin{algorithmic}[1]
\State $ X' \gets X $, $ n \gets 0 $
\While{$ O(X') = 1$ \& $n < n_{max} $ \& $n < F.length $}
    
	\State $ X' \gets AddFeature(X', F[n]) $ %

    \State $ r \gets RandomInteger(0, Length(F)) $  %
	\State $ F_{r} \gets ChooseRandomFeatures(r, F)$
	\State $ X' \gets AddFeatures(X', F_{r}) $
	
	\State $ X' \gets ValidatePerturbations(X, X') $ %
	\State $ n \gets n + 1$
	\If{$ O(X') = 0 $}{
    	\Return $ Success $
    }
    \EndIf
\EndWhile
\State \Return $ Failure $
\end{algorithmic}
\end{algorithm}
    \end{minipage}
  }
\end{figure}

\vspace{-10mm}

\begin{figure}[!htbp]
  \centering
  \resizebox{0.75\linewidth}{!}{%
    \begin{minipage}{\linewidth}
\begin{algorithm}[H]
\caption{Gray-box query attack: for oracle \(O\), malware sample \(X\), vector of sorted benign features \(\vec{s}\), maximum permitted queries \(n_{max}\).}
\label{alg:statisticalqbased}
\textbf{Input:} \(O\), \(X\), \(\vec{s}\), \(n_{max}\)
\begin{algorithmic}[1]
\State $ X' \gets X $, $ n \gets 0 $
\While{$ O(X') = 1$ \& $n < n_{max} $ \& $n < \vec{s}.length $}
	\State $ X' \gets AddFeature(X', \vec{s}[n]) $ %
	\State $ r \gets RandomInteger(0, Length(\vec{s})) $  %
	\State $ F_{r} \gets ChooseRandomFeatures(r, \vec{s})$
	\State $ X' \gets AddFeatures(X', F_{r}) $
	
	\State $ X' \gets ValidatePerturbations(X, X') $ %
	\State $ n \gets n + 1$
	\If{$ O(X') = 0 $}{
    	\Return $ Success $
    }
    \EndIf
\EndWhile
\State \Return $ Failure $
\end{algorithmic}
\end{algorithm}
    \end{minipage}
  }
\end{figure}

\vspace{-10mm}

\begin{figure}[!htbp]
  \centering
  \resizebox{0.75\linewidth}{!}{%
    \begin{minipage}{\linewidth}
\begin{algorithm}[H]
\caption{Adaptive attack: for oracle \(O\), malware sample \(X\), vector of sorted benign features \(\vec{s}\), maximum permitted queries \(n_{max}\), \(p\) percentage of features to remove, \(m\) maximum features to add in each iteration.}
\label{alg:statisticalqbased}
\textbf{Input:} \(O\), \(X\),  \(\vec{s}\), \(n_{max}\), \(m\)
\begin{algorithmic}[1]
\State $ X' \gets X $, $ n \gets 0 $
\While{$ O(X') = 1$ \& $n < n_{max} $ \& $n < \vec{s}.length $}
	\State $ X' \gets AddFeature(X', \vec{s}[n]) $ %
	\State $ r \gets RandomInteger(0, m) $  
	\State $ F_{r} \gets ChooseRandomFeatures(r, \vec{s})$
	\State $ X' \gets AddFeatures(X', F_{r}) $
	
	\State $ X' \gets RemoveFeatures(X', p) $ %
	
	\State $ X' \gets ValidatePerturbations(X, X') $ %
	\State $ n \gets n + 1$
	\If{$ O(X') = 0 $}{
    	\Return $ Success $
    }
    \EndIf
\EndWhile
\State \Return $ Failure $
\end{algorithmic}
\end{algorithm}
    \end{minipage}
  }
\end{figure}

\section{Extended Results}
\label{appendix:extendedresults}
\vspace{-1.5mm}
The extended results are located at: \url{https://osf.io/sfvyn/?view_only=2caaef2fd7ae416a8891ce3f3bd50d2d}

\section{Evaluating Other Query Attacks}
\label{appendix:additionalqueryattacks}

We also examine the Boundary \cite{brendel2017decision}, HopSkipJump \cite{chen2020hopskipjumpattack}, and ZOO \cite{chen2017zoo} query attacks. We demonstrate their inability to generate adversarial examples in this domain, as these attacks are designed for other domains and do not consider the constraints of the malware detection domain (functionality preservation and discretization of features). We use each attack to try and generate adversarial examples against the NN-AT model. For this, once the feature vector of a malware sample has been perturbed, we discretize the feature vectors and evaluate whether the perturbations are permitted for each dataset. As usual, any invalid perturbations are reversed.

\begin{figure}[!htbp]
\vspace{2mm}
	\centering
	\begin{subfigure}[b]{0.15\textwidth}
		\begin{tikzpicture}[scale=0.49]
			\begin{axis}[
				xlabel={\(n_{max}\)},
				ylabel={Evasion rate (\%)}, 
				xmin=0, xmax=500,
				ymin=0, ymax=100,
				xtick={0,100,200,300,400,500},
				ytick={0,20,40,60,80,100},
				ymajorgrids=false,
				legend pos=south west,
				legend style={nodes={scale=0.65, transform shape}},
				cycle list name=color list,
				legend to name=additionalqueryattackslegend,
				legend columns=8,
				height=4cm,
				width=5.9cm, xlabel style = {font=\Large},ylabel style = {font=\large}
			]

			\addplot[color=red]
			coordinates {
				(0,0)(100,77.8)(200,79.9)(300,84.2)(400,86.3)(500,86.3)
			};
			\addlegendentry{Black-box query attack}

			\addplot
			coordinates {
				(0,0)(100,0)(200,0)(300,0)(400,0)(500,0)
			};
			\addlegendentry{Boundary}
			
			\addplot
			coordinates {
				(0,0)(100,0)(200,0)(300,0)(400,0)(500,0)
			};
			\addlegendentry{HopSkipJump}
						   
			\addplot
			coordinates {
				(0,0)(100,0)(200,0)(300,0)(400,0)(500,0)
			};
			\addlegendentry{ZOO}
			\end{axis}
		\end{tikzpicture}
		\caption{AndroZoo}
	\end{subfigure}
	\begin{subfigure}[b]{0.15\textwidth}
		\begin{tikzpicture}[scale=0.49]
			\begin{axis}[
				xlabel={\(n_{max}\)},
				ylabel={Evasion rate (\%)}, 
				xmin=0, xmax=500,
				ymin=0, ymax=100,
				xtick={0,100,200,300,400,500},
				ytick={0,20,40,60,80,100},
				ymajorgrids=false,
				legend pos=south west,
				legend style={nodes={scale=0.65, transform shape}},
				cycle list name=color list,
				legend to name=additionalqueryattackslegend,
				legend columns=8,
				height=4cm,
				width=5.9cm, xlabel style = {font=\Large},ylabel style = {font=\large}
			]

			\addplot[color=red]
			coordinates {
				(0,0)(100,98.69)(200,100)(300,100)(400,100)(500,100)
			};
			\addlegendentry{Black-box query attack}

			\addplot
			coordinates {
				(0,0)(100,0)(200,0)(300,0)(400,0)(500,0)
			};
			\addlegendentry{Boundary}
			
			\addplot
			coordinates {
				(0,0)(100,0)(200,0)(300,0)(400,0)(500,0)
			};
			\addlegendentry{HopSkipJump}
						   
			\addplot
			coordinates {
				(0,0)(100,0)(200,0)(300,0)(400,0)(500,0)
			};
			\addlegendentry{ZOO}
			\end{axis}
		\end{tikzpicture}
		\caption{SLEIPNIR}
	\end{subfigure}
	\begin{subfigure}[b]{0.15\textwidth}
		\begin{tikzpicture}[scale=0.49]
			\begin{axis}[
				xlabel={\(n_{max}\)},
				ylabel={Evasion rate (\%)}, 
				xmin=0, xmax=500,
				ymin=0, ymax=100,
				xtick={0,100,200,300,400,500},
				ytick={0,20,40,60,80,100},
				ymajorgrids=false,
				legend pos=south west,
				legend style={nodes={scale=0.65, transform shape}},
				cycle list name=color list,
				legend to name=additionalqueryattackslegend,
				legend columns=8,
				height=4cm,
				width=5.9cm, xlabel style = {font=\Large},ylabel style = {font=\large}
			]
			\addplot[color=red]
			coordinates {
				(0,0)(100,98.69)(200,100)(300,100)(400,100)(500,100)
			};
			\addlegendentry{Black-box query attack}

			\addplot
			coordinates {
				(0,0)(100,0)(200,0)(300,0)(400,0)(500,0)
			};
			\addlegendentry{Boundary}
			
			\addplot
			coordinates {
				(0,0)(100,0)(200,0)(300,0)(400,0)(500,0)
			};
			\addlegendentry{HopSkipJump}
						   
			\addplot
			coordinates {
				(0,0)(100,0)(200,0)(300,0)(400,0)(500,0)
			};
			\addlegendentry{ZOO}

			\end{axis}
		\end{tikzpicture}
		\caption{DREBIN}
	\end{subfigure}
	\ref{additionalqueryattackslegend}
\vspace{-2mm}
 	\caption{Evasion rate of additional query attacks against NN-AT model vs. \(n_{max}\).}
	\label{figure:additionalqueryattacks}
		
\vspace{-3.5mm}
\end{figure}
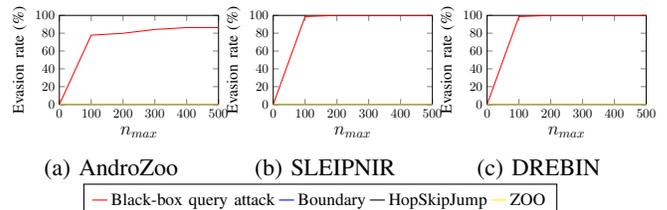

Figure~\ref{figure:additionalqueryattacks} shows that these attacks are unable to achieve evasion \emph{at all}. Evidently, the perturbations used to cross the decision boundary are reversed. That is, invalid perturbations are consistently made that must be discretized and reversed to ensure functionality preservation in the feature-space. Meanwhile, in Figure~\ref{figure:additionalqueryattacks}, we also show that our black-box query attack strategy achieves \(80+\%\) evasion rate across the datasets.
\vspace{-5mm}
\vspace{-3mm}

\end{appendices}

\end{document}